\pdfoutput=1

\documentclass[11pt]{article}

\usepackage[preprint]{acl}

\usepackage{times}
\usepackage{latexsym}

\usepackage[T1]{fontenc}

\usepackage[utf8]{inputenc}

\usepackage{microtype}

\usepackage{inconsolata}

\usepackage[shortlabels]{enumitem}

\usepackage{amsmath}
\usepackage{amssymb}
\usepackage{mathtools}
\usepackage{amsthm}
\usepackage{inconsolata}
\usepackage{longtable}
\usepackage{graphicx}
\usepackage{subcaption}
\usepackage{makecell}
\usepackage{booktabs}
\usepackage{fancyvrb}
\usepackage[safe]{tipa}
\usepackage[capitalize,noabbrev]{cleveref}

\theoremstyle{plain}

\theoremstyle{definition}

\theoremstyle{remark}

\usepackage[textsize=tiny]{todonotes}

\newcommand{\GetVals}{\texttt{GetVals}}
\newcommand{\GetFeature}{\texttt{GetFeature}}
\newcommand{\vals}{\mathsf{Values}}
\newcommand{\CE}{\texttt{CE}}
\newcommand{\loss}{\mathcal{L}}
\newcommand{\II}{\texttt{II}}

\newcommand{\Score}{\texttt{Disentangle}}
\newcommand{\changeScore}{\texttt{Cause}}
\newcommand{\nochangeScore}{\texttt{Iso}}

\newcommand{\model}{\mathcal{M}}

\newcommand{\rep}{\mathbf{N}}
\newcommand{\reps}{\mathcal{N}}
\newcommand{\repval}{\mathbf{n}}
\newcommand{\attributes}{\mathcal{A}}

\newcommand{\attribute}{A}

\newcommand{\entities}{\mathcal{E}}
\newcommand{\entity}{E}
\newcommand{\getattr}[2]{#1_{#2}}
\newcommand{\featurizer}{\mathcal{F}}
\newcommand{\feature}{F}

\newcommand{\featureval}{f}

\newcommand{\predfunc}{\tau}
\newcommand{\dataset}{\mathcal{D}}
\newcommand{\attprompts}[2]{\mathcal{P}_{#2}^{#1}}
\newcommand{\wikiprompts}[1]{\mathcal{W}_{#1}}
\newcommand{\layer}{L}
\newcommand{\token}{t}

\newcommand{\entitytoken}{\token_{\entity}}

\newcommand{\ourdataset}{\textsc{Ravel}}
\newcommand{\Ourdataset}{\textsc{Ravel}}
\newcommand{\ourdatasetfull}{Resolving Attribute--Value Entanglements in Language Models}
\newcommand{\llama}{Llama2}
\newcommand{\entitysplit}{\texttt{Entity}}
\newcommand{\contextsplit}{\texttt{Context}}

\title{\ourdataset: Evaluating Interpretability Methods on\\ Disentangling Language Model Representations}

\author{Jing Huang\\
  Stanford University\\
  \texttt{hij@stanford.edu} \\\And
Zhengxuan Wu\\
  Stanford University\\
  \texttt{wuzhengx@stanford.edu} \\\And
  Christopher Potts\\
  Stanford University \\
  \texttt{cgpotts@stanford.edu} \\\AND
  Mor Geva\\
 Tel Aviv University\\
 \texttt{morgeva@tauex.tau.ac.il}\\
  \And 
    Atticus Geiger\\
    Pr(Ai)$^2$R Group\\
    \texttt{atticusg@gmail.com}\\
  }

\begin{document}
\maketitle

\begin{abstract}
Individual neurons participate in the representation of multiple high-level concepts. To what extent can different interpretability methods successfully disentangle these roles? To help address this question, we introduce \textbf{\ourdataset} (\ourdatasetfull), a dataset that enables tightly controlled, quantitative comparisons between a variety of existing interpretability methods. We use the resulting conceptual framework to define the new method of Multi-task Distributed Alignment Search (MDAS), which allows us to find distributed representations satisfying multiple causal criteria. 
With \llama-7B as the target language model, MDAS achieves state-of-the-art results on \ourdataset, demonstrating the importance of going beyond neuron-level analyses to identify features distributed across activations. We release our benchmark at \url{https://github.com/explanare/ravel}.
\end{abstract}

\section{Introduction}
 A central goal of interpretability is to localize an abstract concept to a component of a deep learning model that is used during inference. However, this is not as simple as identifying a neuron for each concept, because neurons are \textit{polysemantic} -- they represent multiple high-level concepts \citep{Smolensky1988, PDP1, PDP2,olah2020zoom,cammarata2020thread,bolukbasi:2021, Gurnee:2023}. 
 
 \begin{figure}
 \centering
\includegraphics[width=1.0\linewidth,trim={0 25ex 49ex 0},clip]{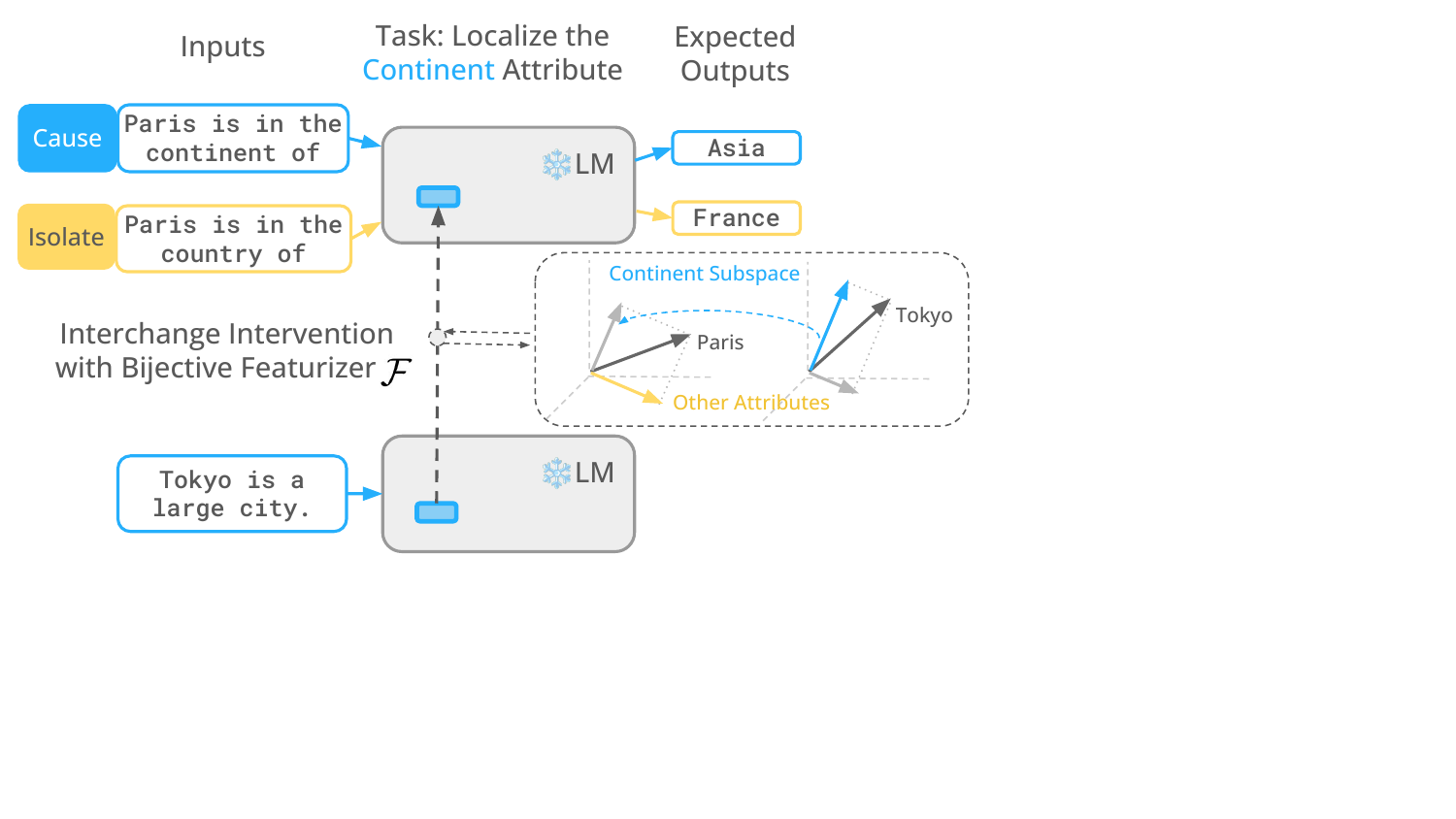}
     \caption{An overview of the \ourdataset\ benchmark, which evaluates how well an interpretability method can find features that isolate the causal effect of individual attributes of an entity.}
     \label{fig:overview}
     \vspace{-3ex}
 \end{figure}

Several recent interpretability works \cite{bricken2023monosemanticity, cunningham2023sparse, geiger2023finding, wu2023} tackle this problem using a \textit{featurizer} that disentangles the activations of polysemantic neurons by mapping to a space of \textit{monosemantic} features that each represent a distinct concept. Intuitively, these methods should have a significant advantage over approaches that identify concepts with sets of neurons. However, these methods have not been benchmarked.
 
To facilitate these method comparisons, we introduce a diagnostic benchmark, \textbf{\ourdataset} (\ourdatasetfull). \ourdataset\ evaluates interpretability methods on their ability to localize and disentangle the attributes of different types of entities encoded as text inputs to language models (LMs). For example, the entity type ``city'' has instances such as ``Paris'' or ``Tokyo'', which each have attributes for ``continent'', namely ``Europe'' and ``Asia''. An interpretability method must localize this attribute to a group of neurons $\rep$, learn a featurizer $\featurizer$ (e.g., a rotation matrix or sparse autoencoder), and identify a feature $\feature$ (e.g., a linear subspace of the residual stream in a Transformer) for the attribute. \Ourdataset\ contains five types of entities (cities, people names, verbs, physical objects, and occupations), each with at least 500 instances, at least 4 attributes, and at least 50 prompt templates per entity type. 

The metric we use to assess interpretability methods is based on interchange interventions (also known as activation patching). This operation has emerged as a workhorse in interpretability, with a wide swath of research applying the technique to test if a high-level concept is stored in a model representation and used during inference \cite{Geiger-etal-2020, vig, geiger2021causal, Li2021, finlayson-2021-causal,meng2022locating, chan2022causal,geva-2023-dissecting, wang2023, hanna2023, Conmy2023, goldowsky2023localizing, hase2023does,Todd2023, Feng2023, cunningham2023sparse, huang2023assess, Tigges2023, Lieberum2023, Davies2023, Hendel2023,ghandeharioun2024patchscopes}.

Specifically, we use the LM to process a prompt like ``\texttt{Paris is in the continent of}'' and then intervene on the neurons $\rep$ to fix the feature $\feature$ to be the value it would have if the LM were given a prompt like ``\texttt{Tokyo is a large city.}'' If this leads the LM to output ``\texttt{Asia}'' instead of ``\texttt{Europe}'', then we have evidence that the feature $\feature$ encodes the attribute ``continent''. Then, we perform the same intervention when the LM processes a prompt like ``\texttt{People in Paris speak}''. If the LM outputs ``\texttt{French}'' rather than ``\texttt{Japanese}', then we have evidence that the feature $\feature$ has disentangled the attributes ``continent'' and ``language''.

A variety of existing interpretability methods are easily cast in the terms needed for \ourdataset\ evaluations, including supervised probes \cite{Peters2018, Hupkes2018, tenney-2019-bert, clark-2019-bert}, Principal Component Analysis \cite{ Tigges2023, marks2023geometry}, Differential Binary Masking (DBM: \citealt{Cao2020, Csordas2021, Cao2022, Davies2023}), sparse autoencoders \cite{bricken2023monosemanticity, cunningham2023sparse}, and Distributed Alignment Search (DAS: \citealt{geiger2023finding, wu2023}). Our apples-to-apples comparisons reveal conceptual similarities between the methods. 

In addition, we propose multi-task training objectives for DBM and DAS. These objectives allow us to find representations satisfying multiple causal criteria, and we show that Multi-task DAS is the most effective of all the methods we evaluate at identifying disentangled features. This contributes to the growing body of evidence that interpretability methods need to identify features that are distributed across neurons.

\section{The \ourdataset\ Dataset}\label{sec:dataset}
The design of \ourdataset\ is motivated by four high-level desiderata for interpretability methods:
\begin{enumerate}[leftmargin=*,topsep=4pt,parsep=1pt, itemsep=1pt]
\item\label{d:faith} 
\textbf{Faithful}: Interpretability methods should accurately represent the model to be explained.
\item\label{d:causal} 
\textbf{Causal}: Interpretability methods should analyze the causal effects of model components on model input--output behaviors.
\item\label{d:gen}
\textbf{Generalizable}: The causal effects of the identified components should generalize to similar inputs that the underlying model makes correct predictions for.
\item\label{d:isolate}
\textbf{Isolating individual concepts}: Interpretability methods should isolate causal effects of individual concepts involved in model behaviors.
\end{enumerate}
The goal of \ourdataset\ is to assess the ability of methods to isolate individual explanatory factors in model representations (desideratum~\ref{d:isolate}), and do so in a way that is faithful to how the target models work (desideratum~\ref{d:faith}). The dataset train/test structure seeks to ensure that methods are evaluated by how well their explanations generalize to new cases (desideratum~\ref{d:gen}), and \ourdataset\ is designed to support intervention-based metrics that assess the extent to which methods have found representations that causally affect the model behavior (desideratum~\ref{d:causal}).

\begin{table}[t]
  \centering
  \resizebox{1\linewidth}{!}{
  \small
  \setlength{\tabcolsep}{8pt}
  \begin{tabular}%
  {p{0.17\linewidth}p{0.42\linewidth}@{}r r}
 \toprule
  \makecell[tl]{Entity \\ Type} & Attributes & \# Entities & \makecell[tl]{\#~Prompt\\Templates} \\ 
 \midrule
City & \makecell[tl]{Country, Language, \\ Latitude, Longitude, \\ Timezone, Continent} & 3552 & 150 \\ \midrule
\makecell[tl]{Nobel \\ Laureate} & \makecell[tl]{Award Year, Birth Year, \\ Country of Birth, Field, \\ Gender} & 928 & 100 \\ \midrule
Verb & \makecell[tl]{Definition, Past Tense, \\ Pronunciation, Singular} & 986 & 60 \\ \midrule
Physical Object & \makecell[tl]{Biological Category, \\ Color, Size, Texture} & 563 &  60 \\ \midrule
Occupation & \makecell[tl]{Duty, Gender Bias,\\ Industry, Work Location} & 799 &  50 \\

 \bottomrule
\end{tabular}
}
 \caption{Types of entities and attributes in \ourdataset.}
 \label{tab:dataset}
 \vspace{-3ex}
\end{table}

\Ourdataset\ is carefully curated as a diagnostic dataset for the attribute disentanglement problem. \ourdataset\ has five types of entity, where every instance has every attribute associated with its type. Table~\ref{tab:dataset} provides an overview of \ourdataset's structure.

\paragraph{The Attribute Disentanglement Task}  We begin with a set of entities $\entities=\{\entity_1, \dots, \entity_n\}$, each with attributes $\attributes =\{\attribute_1, \dots, \attribute_k\}$, where the correct value of $\attribute$ for $\entity$ is given by $\getattr{\attribute}{\entity}$. 
Our interpretability task asks whether we can find a feature $ \feature$ that encodes the attribute $\attribute$ separately from the other attributes $\attributes \setminus \{\attribute\}$. For Transformer-based models \citep{Vaswani-etal:2017}, a feature might be a dimension in a hidden representation of an MLP or a linear subspace of the residual stream. 

We do not know a priori the degree to which it is possible to disentangle a model's representations. However, our benchmark evaluates interpretability methods according to the desiderata given above and so methods will need to be faithful to the model's underlying structure to succeed. In other words, assuming methods are faithful, we can favor methods that achieve more disentanglement.

\subsection{Data Generation}

\paragraph{Selecting Entity Types and Attributes}
We first identify entity types from existing datasets that potentially have thousands of instances (see Appendix~\ref{appx:attribute}), such as cities or famous people.
Moreover, each entity type has multiple attributes with different degrees and types of associations. For example, for attributes related to city, ``country'' entails ``continent'', but not the reverse; ``country'' is predictable from ``timezone'' but non-entailed; and ``latitude'' and ``longitude'' are the least correlated compared with the previous two pairs, but have identical output spaces. These entity types together cover a diverse set of attributes such that predicting the value of the attribute uses factual, linguistic, or commonsense knowledge.

\paragraph{Constructing Prompts} We consider two types of prompts: attribute prompts and entity prompts. Attribute prompts $\attprompts{\attribute}{\entity}$ contain mentions of $E$ and instruct the model to output the attribute value $\getattr{\attribute}{\entity}$. For example, $\entity$ = $\texttt{Paris}$ is an instance of the type ``city'', which has an attribute $\attribute$ = $\texttt{Continent}$ that can be queried with prompts  ``\texttt{Paris is in the continent of}''. Prompts can also be JSON-format, e.g.,  ``\texttt{\{"city": "Paris", "continent":"}'', which reflects how entity--attribute association might be encoded in training data. For each format, we do zero- and few-shot prompting. In addition to attribute prompts, entity prompts $\wikiprompts{\entity}$ contain mentions of the $\entity$, but does not query any $\attribute\in\attributes$. For example, ``\texttt{Tokyo is a large city}''. We sample entity prompts from the Wikipedia corpus.\footnote{We use the \texttt{20220301.en} version pre-processed by HuggingFace at \url{https://huggingface.co/datasets/wikipedia}}

For a set of entities $\entities$ and a set of attributes to disentangle $\attributes$, the full set of prompts is
\[\dataset =  \{x : x \in \attprompts{\attribute}{\entity} \cup \wikiprompts{\entity}, \entity \in \entities, \attribute \in \attributes \}\] 

\paragraph{Generating Splits}

\Ourdataset\ offers two settings, $\entitysplit$ and $\contextsplit$, to evaluate the \emph{generalizability} (desideratum \ref{d:gen}) of an interpretability method across unseen entities and contexts. Each setting has a predefined train/dev/test structure. In $\entitysplit$, for each entity type, we randomly split the entities into 50\%/25\%/25\% for train/dev/test, but use the same set of prompt templates across the three splits. In $\contextsplit$, for each attribute, we randomly split the prompt templates into 50\%/25\%/25\%, but use the same set of entities across the three splits.

\paragraph{Filtering for a Specific Model}

When evaluating interpretability methods that analyze a model $\model$, we generally focus on a subset of the instances where $\model$ correctly predicts the values of the attributes (see Appendix~\ref{appx:dataset_llama2}). This allows us to focus on understanding why models succeed, and it means that we don't have to worry about how methods might have different biases for incorrect predictions.

\subsection{Interpretability Evaluation}

\paragraph{Interchange Interventions}
A central goal of \ourdataset\ is to assess methods by the extent to which they provide causal explanations of model behaviors (desideratum~\ref{d:causal}). To build such analyses, we need to put models into counterfactual states that allow us to isolate the causal effects of interest.

The fundamental operation for achieving this is the intervention \citep{Spirtes, Pearl:2001, Pearl2009}: we change the value of a model-internal state and study the effects this has on the model's input--output behavior. In more detail: let $\model(x)$ be the entire state of the model when $\model$ receives input $x$, i.e., the set of all input, hidden, and output representations created during inference. Let $\model_{\rep \gets \repval}$ be the model where neurons $\rep$ are intervened upon and fixed to take on the value $\repval \in \vals(\rep)$.

\citet{geiger2023finding} generalize this operation to intervene upon features that are distributed across neurons using a bijective featurizer $\featurizer$. Let $\model_{\feature \gets \featureval}$ be the model where neurons $\rep$ are projected into a feature space using $\featurizer$, the feature $\feature$ is fixed to take on value $\featureval$, and then the features are projected back into the space of neural activations using $\featurizer^{-1}$. If we let $\predfunc(\model(x))$ be the token that a model predicts for a given prompt $x \in \dataset{}{}$, then comparisons between $\predfunc(\model(x))$ and $\predfunc(\model_{\feature \gets \featureval}(x))$ yield insights into the causal role that $\feature$ plays in model behavior.

However, most conceivable interventions fix model representations to be values that are never realized by any input. To characterize the high-level conceptual role of a model representation, we need a data-driven intervention that sets a representation to values it could actually take on. This is achieved by the \emph{interchange intervention}, which fixes a feature $\feature$ to the value it would take if a different input $x'$ were provided:
\begin{multline}
\II(\model, \feature,x,  x') \overset{\text{def}}{=}\\
\predfunc\Big(\model_{\feature \gets \GetFeature(\model(x'), \feature)}(x)\Big)
\end{multline}
where $\GetFeature(\model(x'), \feature)$ is the value of $\feature$ in $\model(x')$.
Interchange interventions represent a very general technique for identifying abstract causal processes that occur in complex black-box systems \citep{Beckers_Halpern_2019,pmlr-v115-beckers20a,Geiger-etal:2023:CA}. 

\paragraph{Evaluation Data}
For evaluation, each intervention example consists of a tuple: an input $x \in \attprompts{\attribute}{\entity}$, an input $x' \in \attprompts{\attribute'}{\entity'} \cup \wikiprompts{\entity'}$, a target attribute $\attribute^*$, and an intervention label $y$. If $\attribute^* = \attribute$, then $y$ is $\getattr{\attribute}{\entity'}$ and otherwise $y$ is $\getattr{\attribute}{\entity}$. 
For example, if the set of ``city'' entities to evaluate on is $\{\texttt{"Paris"}, \texttt{"Tokyo"}\}$ and the goal is to disentangle the ``country'' attribute from the ``continent'' attribute, then the set of test examples becomes the one shown in Figure~\ref{fig:overview}.

\paragraph{Metrics}
If $\model$ achieves behavioral success on a dataset, we can use that dataset to evaluate an interpretability method on its ability to identify a collection of neurons $\rep$, a featurizer $\featurizer$ for those neurons, and a feature $\feature$ that represents an attribute $\attribute$ separately from all others attributes $\attributes \setminus \{\attribute\}$. 

If $\feature$ encodes $\attribute$, then interventions on $\feature$ should change the value of $\attribute$. When $\model$ is given a prompt $x \in \attprompts{\attribute}{\entity}$, we can intervene on $\feature$ to set the value to what it would be if a second prompt $x' \in \attprompts{\attribute'}{\entity'} \cup \wikiprompts{\entity'}$ were provided. The token predicted by $\model$ should change from $\getattr{\attribute}{\entity}$ to $\getattr{\attribute}{\entity'}$:
\begin{multline*}
\changeScore(\attribute, \feature, \model, \dataset) \overset{\text{def}}{=}\\
\mathop{\mathbb{E}_{\dataset}}\left[\II(\model,\feature,x, x') = \getattr{\attribute}{\entity'}\right]
\end{multline*}
If $\feature$ isolates $\attribute$, then interventions on $\feature$ should not cause the values of other attributes $\attribute^* \in \attributes \setminus \{\attribute\}$ to change. When $\model$ is given a prompt $x^* \in \attprompts{\attribute^*}{\entity}$, we can again intervene on $\feature$ to set the value to what it would be if a second prompt $x' \in \attprompts{\attribute'}{\entity'} \cup \wikiprompts{\entity'}$ were provided. The token predicted by $\model$ should remain $\getattr{\attribute^*}{\entity}$:
\begin{multline*}
\nochangeScore(\attribute, \feature, \model, \dataset) \overset{\text{def}}{=}\\
\frac{1}{|\attributes \setminus \{\attribute\}|}\! \sum_{\attribute^* \in \attributes \setminus \{\attribute\}}\!\!\!\!\!
\mathop{\mathbb{E}_{\dataset}}\left[\II(\model, \feature,x^*, x') = \getattr{\attribute^*}{\entity}\right]
\end{multline*}

To balance these two objectives, we define the \Score\ score as a weighted average between \changeScore\ and \nochangeScore.
\begin{multline*}
\Score(\attribute,\feature, \model, \dataset) =\\
\frac{1}{2}\Big[\changeScore(\attribute,\feature, \model, \dataset) + 
\nochangeScore(\attribute, \feature, \model, \dataset)\Big]
\end{multline*}

The score on \ourdataset\ for an entity type is its average \Score\ score over all attributes.

In practice, two attributes might not be fully disentanglable in the model $\model$ so there is no guarantee that it is possible to find a feature $\feature$ that achieves $\changeScore=1$ and $\nochangeScore=1$ at the same time. However, evidence that two attributes might not be separable is an insight into how knowledge is structured in the model.

\section{Interpretability Methods}
\label{sec:methods}
We use \ourdataset\ to evaluate a variety of interpretability methods on their ability to disentangle attributes while generalizing to novel templates and entities. Each method uses data from the training split to find a set of neurons $\rep$, learn a featurizer $\featurizer$, and find a feature $\feature_{\attribute}$ that captures an attribute $\attribute \in \attributes$ independent from the other attributes. In Section~\ref{sec:results}, we describe the baseline procedure we use for considering different sets of neurons. In this section, we define methods for learning a featurizer and identifying a feature given a set of neurons. For each method, the core intervention for $\attribute$ is given by $\II(\model, \feature_{\attribute},  x, x')$ where $\feature_{\attribute}$ is defined by the method. In this section, we use $\GetVals(\model(x), \rep)$ to mean the activations of neurons $\rep$ when $\model$ processes input $x$.

\subsection{PCA}
Principal Component Analysis (PCA) is a dimensionality reduction method that minimizes information loss. In particular, given a set of real valued vectors $\mathcal{V} \subset \mathbb{R}^{n}$, $|\mathcal{V}|>n$, the principal components are $n$ orthogonal unit vectors $\mathbf{p}_1, \dots, \mathbf{p}_{n}$ that form an $n \times n$ matrix:
\[\mathsf{PCA}(\mathcal{V}) = \begin{bmatrix} \mathbf{p}_1 & \dots & \mathbf{p}_{n}\end{bmatrix}\]
For our purposes, the orthogonal matrix formed by the principal components serves as a featurizer that maps neurons $\rep$ into a more interpretable space \cite{chormai2022disentangled, marks2023geometry, Tigges2023}. Given an attribute $A$, a training dataset $\dataset$ from \ourdataset\ for a particular entity type, a model $\model$, and a set of neurons $\rep$, we define
\begin{multline*}
\featurizer_{\attribute}(\repval) =\\
\repval^{T}\mathsf{PCA}(\{\GetVals(\model(x), \rep): x \in \dataset\}) 
\end{multline*}

PCA is an unsupervised method, so there is no easy way to tell what information is encoded in each principal component. To solve this issue, for each attribute $\attribute \in \attributes$ we train a linear classifier with L1 regularization to predict the value of $\attribute$ from the featurized neural representations. Then, we define the feature $\feature_{\attribute}$ to be the set of dimensions assigned a weight by the classifier that is greater
than a hyperparameter $\epsilon$.

\subsection{Sparse Autoencoder}
A recent approach to featurization is to train an autoencoder to project neural activations into a higher dimensional sparse feature space and then reconstruct the neural activations from the features \cite{bricken2023monosemanticity, cunningham2023sparse}. We train a sparse autoencoder on the loss
\[\sum_{x \in \dataset}||\GetVals(\model(x), \rep) - \Big(W_2\mathbf{f} + b_2 \Big)||_2 + ||\mathbf{f}||_{1}\]
 \[\mathbf{f} = \mathsf{ReLU}(W_1(\GetVals(\model(x), \rep)-b_2) + b_1)\] with $W_1 \in \mathbb{R}^{n \times m}$, $W_2 \in \mathbb{R}^{m \times n}$, $b_1 \in \mathbb{R}^m$, and $b_2 \in \mathbb{R}^n$.
To construct a training datset, we sample 100k sentences from the Wikipedia corpus for each entity type, each containing a mention of an entity in the training set. We extract the 4096-dimension hidden states of \llama-7B at the target intervention site as the input for training a sparse autoencoder with 16384 features.
 
 We use the autoencoder to define a featurizer
\[\featurizer_{\attribute}(\repval) = \mathsf{ReLU}(W_1(\repval - b_2) + b_1) \]
and an inverse $\featurizer^{-1}_{\attribute}(\repval) = W_2\repval + b_2$.

An important caveat to this method is that the featurizer is only truly invertible if the autoencoder has a reconstruction loss of 0. The larger the loss is, the more unfaithful this interpretability method is to the model being analyzed. All other methods considered use an orthogonal matrix, which is truly invertible up to floating point precision.

Similar to PCA, sparse autoencoders are an unsupervised method that does not produce features with obvious meanings. Again, to solve this issue, for each attribute $\attribute \in \attributes$ we train a linear classifier with L1 regularization and define the feature $\feature_{\attribute}$ to be the set of dimensions assigned a weight that is greater than a hyperparameter $\epsilon$.

\subsection{Relaxed Linear Adversarial Probe}
Supervised probes are a popular interpretability technique for analyzing how neural activations correlate with high-level concepts \cite{Peters2018, Hupkes2018, tenney-2019-bert, clark-2019-bert}. When probes are arbitrarily powerful, this method is equivalent to measuring the mutual information between the neurons and the concept \cite{pimentel-2020-information,hewitt-2021-conditional}. However, probes are typically simple linear models in order to capture how easily the information about a concept can be extracted. Probes have also been used to great effect on the task of concept erasure \cite{ravfogel-2020-null,elazar-2021-amnesic, pmlr-v162-ravfogel22a}.

Following the method of \citet{pmlr-v162-ravfogel22a}, we train a relaxed linear adversarial probe (RLAP)  to learn a linear subspace parameterized by a set of $k$ orthonormal vectors $W \in \mathbb{R}^{k \times n}$ that captures an attribute $\attribute$, using the following loss objective:
\[\min_{\theta} \max_{W}\!\! \sum_{x \in \dataset}\! \mathsf{CE}\big (\theta^{T} \mathbf{f}, \attribute_{\entity_x}\big) \]
\[\mathbf{f}=(I - W^{T}W)\big(\GetVals(\model(x), \rep)\big)\]
where $\mathbf{f}$ is the representation of the entity with the attribute information erased, and $\theta$ is a linear classifier that tries to predict the attribute value $\attribute_{\entity_x}$ from the erased entity representation.

We define the $\featurizer$ using the set of $k$ orthonormal vectors that span the row space of $W$ and the set of $n-k$ orthonormal vectors that span the null space:
\[\featurizer_{\attribute}(\repval) = \repval \begin{bmatrix}
\mathbf{r}_1 & \dots & \mathbf{r}_k & \mathbf{u}_{k+1} & \dots & \mathbf{u}_{n}
\end{bmatrix}\]

Our feature $\feature_{\attribute}$ is the first $k$ dimensions of the feature space, i.e. the row space of $W$.
Intuitively, since the linear probe was trained to extract the attribute $\attribute$, the rowspace is the linear subspace of neural activations that the probe is ``looking at'' to make predictions.

\subsection{Differential Binary Masking}

Differential Binary Masking (DBM) learns a binary mask to select a set of neurons that causally represents a concept \cite{Cao2020, Csordas2021, Cao2022, Davies2023}. 
The loss objective used to train the mask is a combination of matching the counterfactual behavior and forcing the mask to be sparse with coefficient $\lambda$:
\begin{multline*}
\loss_{\changeScore} = \CE (\tau(\model_{\rep \gets \repval}(x)), \getattr{\attribute}{\entity'}) + \lambda ||\mathbf{m}||_{1}\\
\repval=  \big(\mathbf{1}- \sigma(\textbf{m} /T)\big)\circ \GetVals(\model(x), \rep) \\
+ \sigma(\textbf{m} /T) \circ \GetVals(\model(x'), \rep)
\end{multline*}
where the intervention is determined by inputs $x,x'$ and learnable parameter $\mathbf{m} \in \mathbb{R}^n$, where $\circ$ is element wise multiplication and $T \in \mathbb{R}$ is a temperature annealed throughout training.

The feature space is the original space of neural activations, i.e., featurizer $\featurizer_{\attribute}(\repval)  = \repval$. The feature $\feature_{\attribute}$ is the set of dimensions $i$ where $1 - \sigma(\textbf{m}_{i} /T) < \epsilon$ for a (small) hyperparameter $\epsilon$.

\begin{table}[t]
  \centering
  \small
  \begin{tabular}[b]{lccc}
 \toprule
  Method & Supervision & $\entitysplit$ & $\contextsplit$ \\ 
 \midrule
Full Rep. & None & 40.5 & 39.5 \\
PCA & None & 39.5 & 39.1 \\
SAE  & None  &  48.6 & 46.8 \\
RLAP & Attribute  & 48.8 & 50.9 \\
DBM & Counterfactual & 52.2  & 49.8 \\
DAS & Counterfactual & 56.5 & 57.3 \\
MDBM & Counterfactual & 53.7 & 53.9 \\
MDAS & Counterfactual & \textbf{60.1} & \textbf{65.6} \\
 \bottomrule
\end{tabular}
 \caption{The disentanglement score on \ourdataset\ for each interpretability method. Numbers are represented in \%.}
 \label{tab:results_all_methods}
 \vspace{-3ex}
\end{table}

\subsection{Distributed Alignment Search}
Distributed Alignment Search (DAS) \cite{geiger2023finding} learns a linear subspace of a model representation with a training objective defined using interchange interventions. In the original work, the linear subspace learned by DAS is parameterized as an $n\times n$ orthogonal matrix $Q = [\mathbf{u}_{1} \dots \mathbf{u}_{n}]$, which rotates the representation into a new coordinate system, i.e., $\featurizer_{\attribute}(\repval)  = Q^{\top} \repval$. The set of feature $\feature_{\attribute}$ is the first $k$ dimensions of the rotated subspace, where $k$ is a hyperparameter. The matrix $Q$ is learned by minimizing the following loss:
 \[\loss_{\changeScore}(\attribute, \feature_{\attribute}, \model) = \CE (\II(\model, \feature_{\attribute},x, x'), \getattr{\attribute}{\entity'})\]
Computing $Q$ is expensive, as it requires computing $n$ orthogonal vectors. To avoid instantiating the full rotation matrix, we use an alternative parameterization where we only learn the $k \ll n$ orthogonal vectors that form the feature $\feature_{\attribute}$ (see Appendix~\ref{appx:das}).

\subsection{Multi-task DBM and DAS}

To address the disentanglement problem, we propose a multitask extension of DBM (MDBM) and DAS (MDAS). The original training objective of DBM and DAS only optimizes for the $\changeScore$ score, without considering the impact on the $\nochangeScore$ score. We introduce the $\nochangeScore$ aspect into the training objective through multitask learning. For each attribute $\attribute^{*}\in \attributes \setminus \{\attribute\}$, we define the $\nochangeScore$ objective as
 \[\loss_{\nochangeScore}(\attribute, \feature_{\attribute}, \model) = \CE (\II(\model(x), \feature_{\attribute}, x'), \getattr{\attribute^{*}}{\entity})\]
We minimize a linear combination of losses from each task:
\begin{multline*}
    \mathcal{L}_{\Score}(\attributes, \feature_{\attribute}, \model) =\\ \loss_{\changeScore}(\attribute, \feature_{\attribute}, \model) + 
    \sum_{A^{*} \in \attributes \setminus \{\attribute\}} \frac{\loss_{\nochangeScore}(\attribute^{*}, \feature_{\attribute}, \model)}{|\attributes \setminus \{\attribute\}|}
\end{multline*}

 \begin{figure}
     \centering
    \includegraphics[width=\linewidth,angle=0]{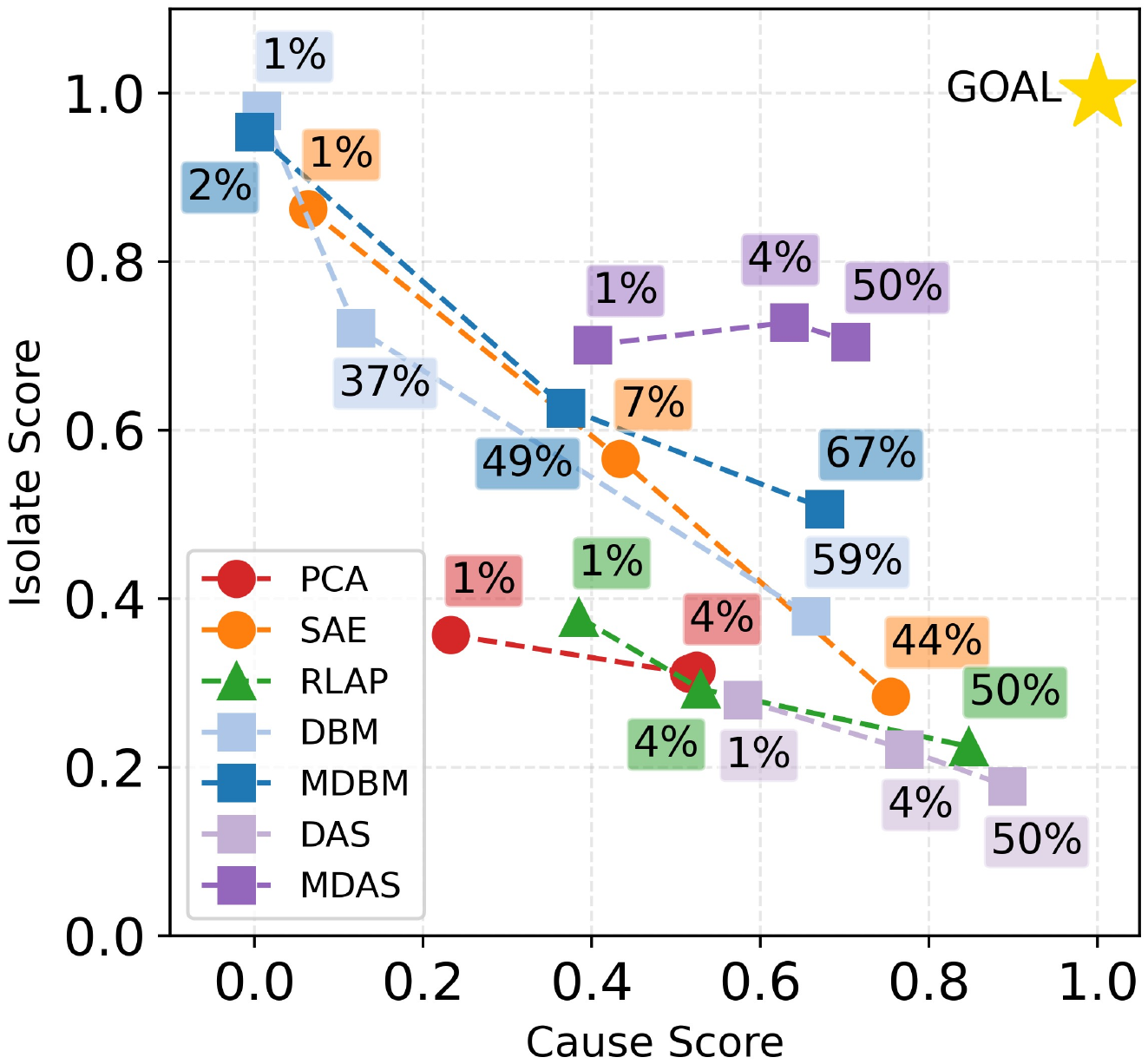}
      \caption{$\changeScore$ and $\nochangeScore$ scores for each method when using different feature sizes, shown as the ratio (\textbf{\%}) between the dimension of $\feature_{\attribute}$ and the dimension of the output space of $\featurizer$. Each method has three data points that vary from using very few ($\approx$1\%) to half ($\approx$50\%) of the dimensions. Increasing feature dimensions generally leads to higher $\changeScore$ score, but lower $\nochangeScore$ score. Figure best viewed in color.}
     \label{fig:result_inv_dim}
             \vspace{-3ex}
 \end{figure}

\section{Experiments}\label{sec:results}
We evaluate the methods described in Section~\ref{sec:methods} on \ourdataset\ with \mbox{\llama-7B}~\cite{touvron2023llama},
a $32$-layer decoder-only Transformer model, as the target LM. Implementation details of each method are provided in Appendix~\ref{appx:method}.

\begin{figure*}[t]
     \centering
    \begin{subfigure}[t]{0.63\textwidth}
        \centering
        \includegraphics[height=30ex,trim={0 0 5cm 0},clip]{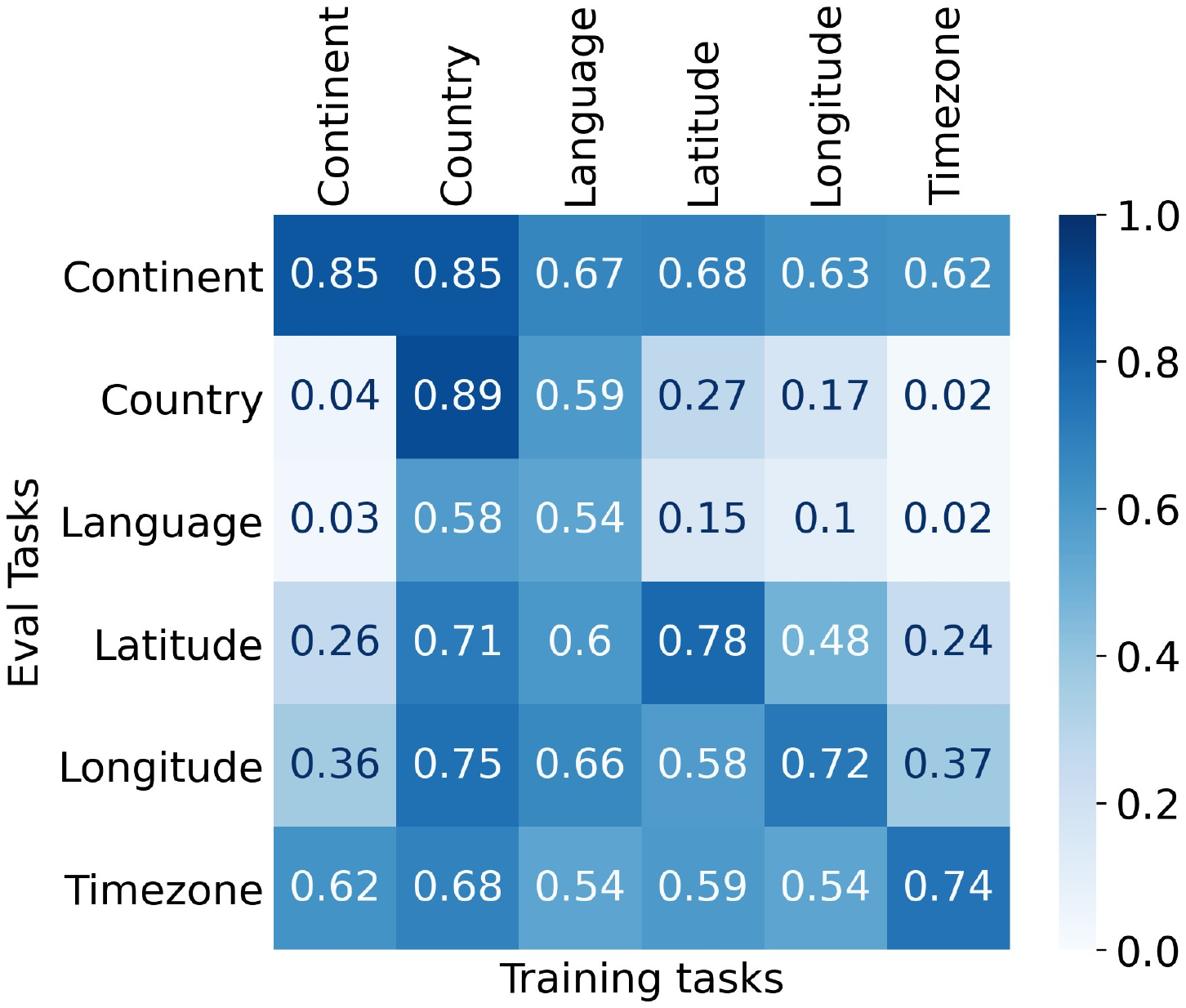}%
        \hfill
         \includegraphics[height=30ex, trim={7.15cm 0 5cm 0},clip]{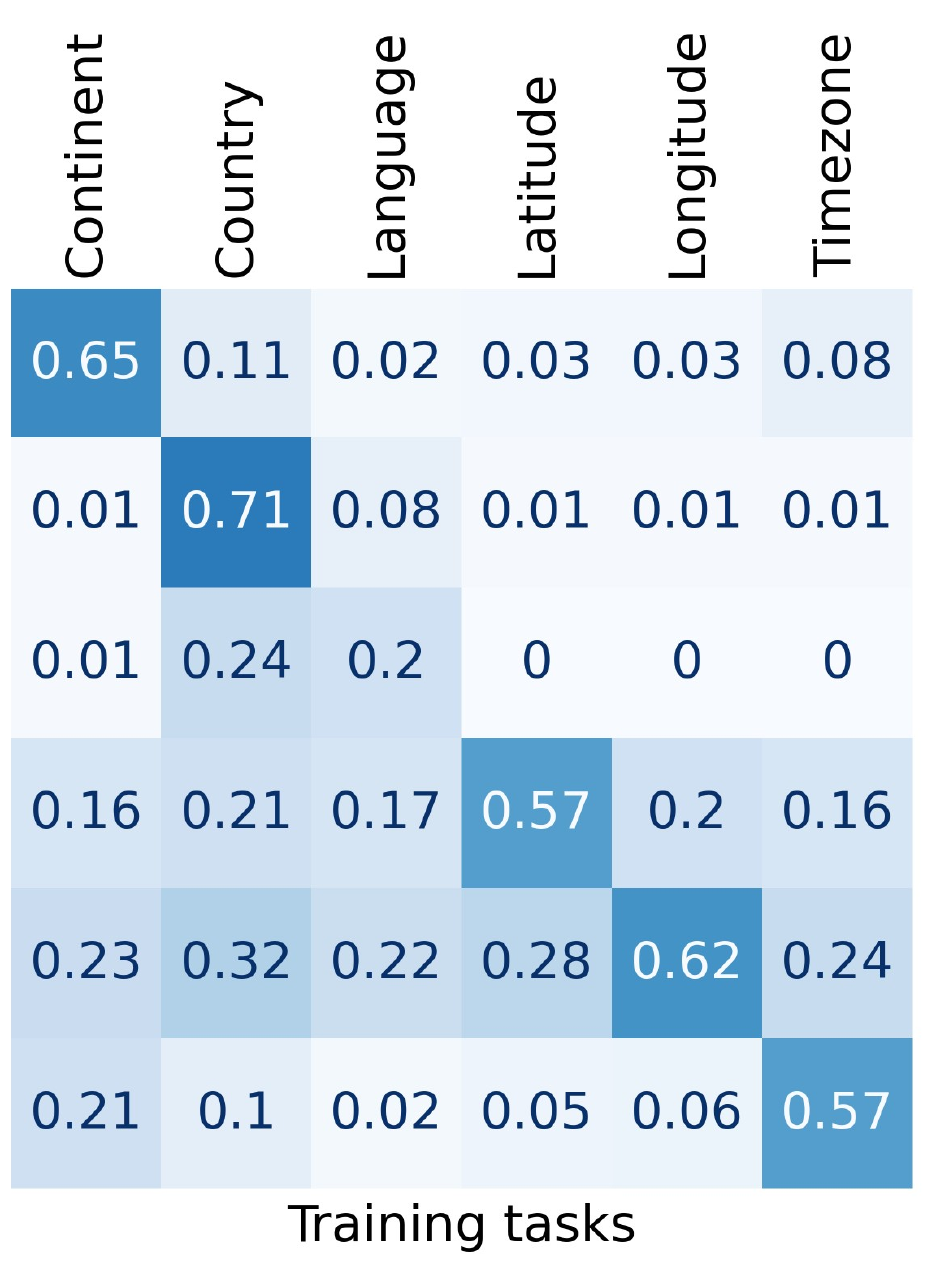}
        \caption{\changeScore\ score for all attributes when intervening on the attribute features identified by DAS (left) and MDAS (right). A \changeScore\ score of 0.62 for column Continent, row Timezone (bottom left corner), means that, when intervening on the Continent feature, the same subspace changes Timezone 62\% of the time.}  \label{fig:attribute_entanglement}
    \end{subfigure}
    \hfill
    \begin{subfigure}[t]{0.35\textwidth}
     \includegraphics[width=\textwidth,angle=0]{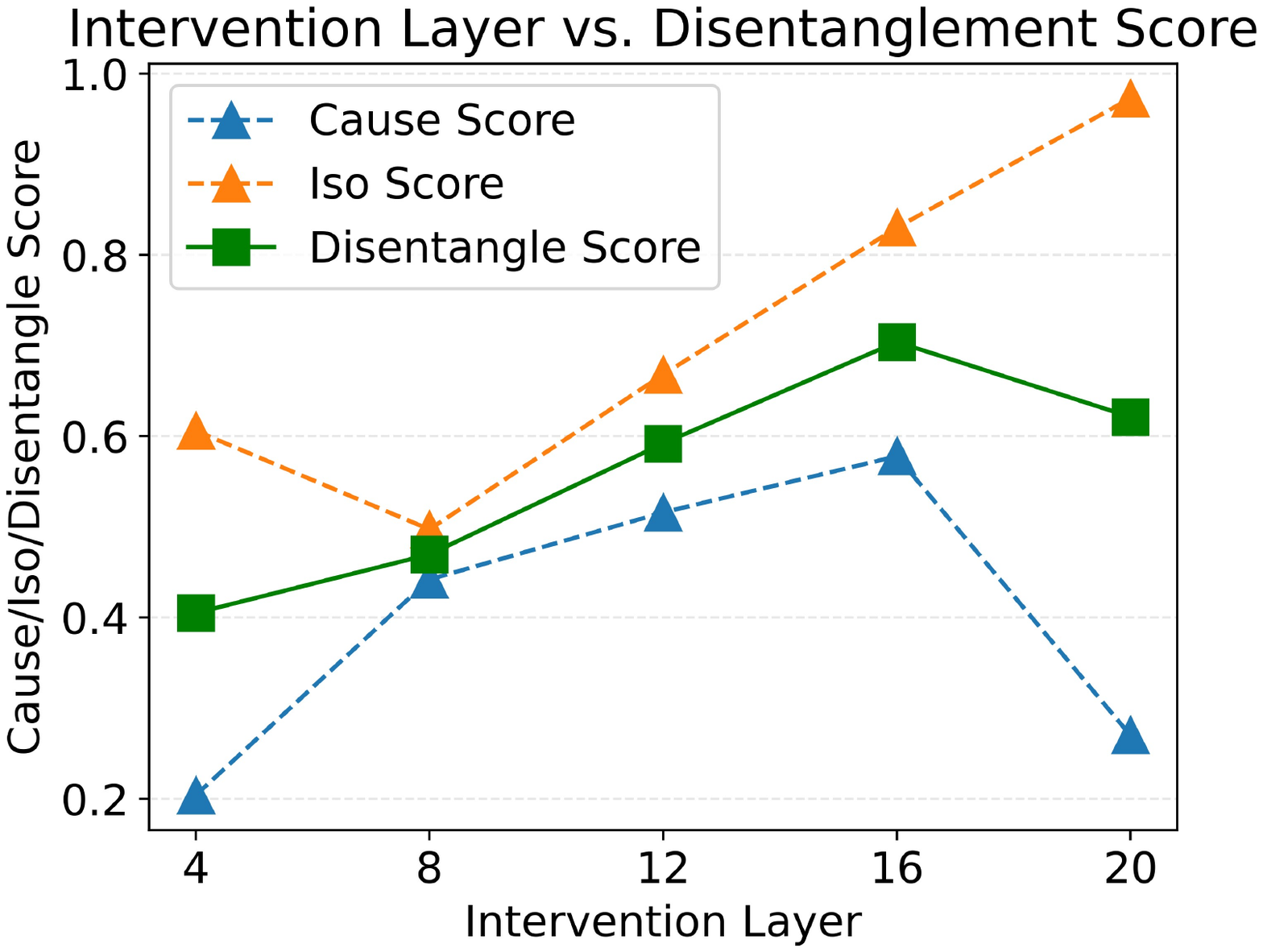}
     \caption{The \changeScore, \nochangeScore, and \Score\ score on the $\entitysplit$ split for the ``country''  feature found by MDAS. The attributes of cities become more disentangled across layers.}
    \label{fig:acrosslayers}
    \end{subfigure}
    \caption{Additional results for the MDAS method.}
    \label{fig:moreresults}
\end{figure*}

\subsection{Setup}

We consider the residual stream representations at the last token of the entity as our potential intervention sites. For autoregressive LMs, the last token of the entity $\entitytoken$ (e.g., the token ``is'' in the case of ``Paris'') likely aggregates information of the entity \cite{meng2022locating,geva-2023-dissecting}. For Transformer-based LMs like \llama, an activation vector $\rep_{\token}^{\layer}$ in the residual stream is created for each token $\token$ at each Transformer layer $\layer$. As the contributions of the MLP and attention heads must pass through the residual stream, it serves as a bottleneck. Therefore, we will limit our methods to examining the set of representations $\reps= \{\rep^{\layer}_{\entitytoken}: \layer \in \{1,\dots,32\}\}$. 

This simplification is only to establish baseline results on the \ourdataset\ benchmark. We expect the best methods will consider other token representations, such as the remainder of the token sequence that realizes the entity.

\subsection{Results}

We evaluate each method on every representation $\rep^{\layer}_{\entitytoken}$ and report the highest disentanglement score on test splits in Table~\ref{tab:results_all_methods}. We additionally include a baseline that simply replaces the full representation $\rep^{\layer}_{\entitytoken}$ regardless of what attribute is being targetted (see Full Rep. in Table~\ref{tab:results_all_methods}). A breakdown of the results with per-attribute \changeScore\ and \nochangeScore\ is in Appendix~\ref{sec:results_appx}.

In Figure~\ref{fig:result_inv_dim}, we show for each method, how the $\nochangeScore$ and $\changeScore$ scores vary as we change the dimensionality of $\feature_{\attribute}$, the feature targeted for intervention. For RLAP, DAS, and MDAS, the dimensionality of $\feature_{\attribute}$ is a hyperparameter we vary directly. For other methods, we vary the coefficient of L1 penalty to vary the size of $\feature_{\attribute}$. Details are given in Appendix~\ref{appx:method}.

In Figure~\ref{fig:moreresults}, we focus on using MDAS, the best performing method, to understand how attributes are disentangled in \llama-7B. Figure~\ref{fig:attribute_entanglement} shows two heat maps summarizing the performance of DAS and MDAS on the entity type ``city''.  These heat maps also show how attributes have different levels of disentanglement. Figure~\ref{fig:acrosslayers} shows how the $\changeScore$, $\nochangeScore$, and $\Score$ scores change for the ``country'' attribute across model layers.

\paragraph{Methods with counterfactual supervision achieve strong results while methods with unsupervised featurizers struggle.}
MDAS is the state-of-the-art method on \ourdataset, being able to achieve high \Score\ scores while only intervening on a feature $\feature_{\attribute}$ with a dimensionality that is $4\%$ of $ |\rep|$ where $\rep$ are the neurons the feature is distributed across (Figure~\ref{fig:result_inv_dim}). DBM, MDBM, and DAS, the other methods that are trained with interventions using counterfactual labels as supervision, achieve the next best performance. PCA and Sparse Autoencoder achieve the lowest \Score\ scores, which aligns with the prior finding that disentangled representations are difficult to learn without supervision \cite{locatello2018challenging}.  Unsurprisingly, more supervision results in higher performance.

\paragraph{Multi-task supervision is better at isolating attributes.} Adding multitask objectives to DBM and DAS increases the overall disentanglement score by 1.5\%/4.1\% and 3.6\%/8.3\% on the $\entitysplit$/$\contextsplit$ split respectively. To further illustrate the differences, we compare DAS with MDAS in Figure~\ref{fig:attribute_entanglement}. On the left, attributes such as ``continent'' and ``timezone'' are naturally entangled with all other attributes; intervening on the feature learned by DAS for any city attribute will also change these two attributes. In contrast, in Figure~\ref{fig:attribute_entanglement} right, MDAS is far more successful at disentangling these attributes, having  small $\changeScore$ scores in all off-diagonal entries.
 
\paragraph{Some groups of attributes are more difficult to disentangle than others.} 
As show in Figure~\ref{fig:attribute_entanglement}, the attribute pairs ``country--language'' and ``latitude--longitude'' are difficult to disentangle. When we train DAS to find a feature for either of these attributes (Figure~\ref{fig:attribute_entanglement} left), the same feature also has causal effects on the other attribute. Even with the additional supervision (Figure~\ref{fig:attribute_entanglement} right), MDAS cannot isolate these attributes. Changing one of these entangled attributes has seemingly unavoidable ripple effects \cite{cohen2023evaluating} that change the other. In contrast, the attribute pair ``language--continent'' can be disentangled. Moreover, the pairs that are difficult to disentangle are consistent across all five supervised methods in our experiment, despite these methods using different training objectives. We include additional visualizations in Appendix~\ref{sec:more_heatmap}. 

\paragraph{Attributes are gradually disentangled across layers.}
The representations of different attributes gradually disentangle as we move towards later layers, as shown in Figure~\ref{fig:acrosslayers}. Early layer features identified by MDAS fail to generalize to unseen entities, hence low $\changeScore$ score. While MDAS is able to identify a feature with relatively high $\changeScore$ starting at layer 8, the $\nochangeScore$ increases from 0.5 to 0.8 from layer 8 to layer 16. It is not until layer 16 that the highest $\Score$ score is achieved.

\section{Related Work}

\paragraph{Intervention-based Interpretability Methods}

Intervention-based techniques, branching off from interchange intervention~\citep{vig, Geiger-etal-2020} or activation patching~\cite{meng2022locating}, have shown promising results in uncovering causal mechanisms of LMs. They play important roles in recent interpretability research of LMs such as causal abstraction~\cite{geiger2021causal, geiger2023finding}, causal tracing to locate factual knowledge~\cite{meng2022locating, geva-2023-dissecting}, path patching or causal scrubbing to find causal circuits~\cite{chan2022causal, Conmy2023, goldowsky2023localizing}, and Distributed Alignment Search~\cite{geiger2023finding}. Previous works suggest that activation interventions that result in systematic counterfactual behaviors provide clear causal insights into model components.

\paragraph{Isolating Individual Concepts}

LMs learn highly distributed representations that encode multiple concepts in a overlapping sets of neurons \cite{Smolensky1988,olah2020zoom,elhage2022toy}. Various methods have been proposed to find components that capture a concept, such as finding a linear subspace that modifies a concept \cite{ravfogel-2020-null,pmlr-v162-ravfogel22a,belrose2023leace,Cao2020,geiger2023finding} and generating a sparse feature space where each direction captures a word sense or is more interpretable \cite{arora-2018-linear,bricken2023monosemanticity,cunningham2023sparse,tamkin2023codebook}. However, these methods have not been evaluated against each other on their ability to isolate concepts. Isolating an individual concept is also related to the goal of ``disentanglement'' in representation learning \cite{RepresentationLearning}, where each direction captures a single generative factor. In this work, we focus on isolating the causal effect of a representation.

\paragraph{Knowledge Representation in LMs}

Understanding knowledge representation in LMs starts with probing structured linguistic knowledge \citep{conneau-2018-cram,tenney-2019-bert,Manning-etal:2020}. Recent work expands to factual knowledge stored in Transformer MLP layers \cite{geva-2021-transformer,dai-2022-knowledge,meng2022locating},  associations represented in linear structures \cite{merullo2023mechanism,hernandez2023linearity,park2023linear}, and deeper study of the semantic enrichment of subject representation \cite{geva-2023-dissecting}. These findings suggest LMs store knowledge modularly, motivating the disentanglement objective in our work.

\paragraph{Benchmarking Interpretability Methods}

Testing the faithfulness of interpretability method relies on counterfactuals. Existing counterfactual benchmarks use behavioral testing \cite{atanasova-2023-faithfulness,schwettmann2023find,mills2023almanacs}, interventions \citep{abraham-etal-2022-cebab}, or a combination of both \citep{huang2023assess}. Recent model editing benchmarks \cite{meng2022locating,zhong-2023-mquake,cohen2023evaluating} also provide counterfactuals that have potential for evaluating interpretability methods.  \textsc{MQuAKE} \cite{zhong-2023-mquake} and \textsc{RippleEdits} \cite{cohen2023evaluating}, in particular, consider entailment relationships of attributes, while we focus on disentanglement.

\section{Conclusion}

We present \ourdataset\, a benchmark for evaluating the ability of interpretability methods to localize and disentangle entity attributes in LMs in a causal,  generalizable manner. We show how \ourdataset\ can be used to evaluate five different families of interpretability methods that are commonly used in the community. We benchmark several strong interpretability methods on \ourdataset\ with \llama{}-7B model as baselines,
and we introduce a multi-task objective that improves the performance of Differential Binary Masking (DBM) and Distributed Alignment Search (DAS). Multi-task DAS achieves the best results in our experiments. Results on our attribute disentanglement task also offer insights into the different levels of entanglement between attributes and the emergence of disentangled representations across layers in the \llama{}-7B model.

The community has seen an outpouring of innovative new interpretability methods. However, these methods have not been systematically evaluated for whether they are \emph{faithful}, \emph{generalizable}, \emph{causally effective}, and able to \emph{isolate individual concepts}. We release \ourdataset\footnote{\url{https://github.com/explanare/ravel}} to the community and hope it will help drive the assessment and development of interpretability methods that satisfy these criteria.

\section*{Limitations}
Our attribute disentanglement results in Section~\ref{sec:results} are based on the \llama-7B model. While \llama-7B uses the widely adopted decoder-only Transformer architecture, different model architectures or training paradigms could produce LMs that favor different interpretability methods. Hence, when deciding which interpretability method is the best to apply to a new model, we encourage people to instantiate $\ourdataset$ on the new model.

When choosing intervention sites, we limit our search to the residual stream above the last entity token. However, representations of attributes can be distributed across multiple tokens or layers. We encourage future work to explore different intervention sites when using this benchmark.

\section*{Ethics Statement}
In this paper, we present an interpretability benchmark that aims to assess the faithfulness, generalizability, causal effects, and the ability to isolate individual concepts in language models. While an interpretability method that satisfies these criteria could be useful for assessing model bias or steering model behaviors, the same method might also be used for manipulating models in undesirable applications such as triggering toxic outputs. These interpretability methods should be studied and used in a responsible manner. 

\section*{Acknowledgements}
This research is supported in part by grants from Open Philanthropy and the Stanford Institute for Human-Centered Artificial Intelligence (HAI).

\bibliography{custom,anthology,2023.emnlp-main}

\newpage
\appendix
\onecolumn

\section{Dataset Details}
\label{appx:dataset}

\begin{table*}[th!]
  \centering
  \small
  \begin{tabular}{p{0.07\linewidth}p{0.03\linewidth}p{0.2\linewidth}p{0.6\linewidth}}
 \toprule
  Attributes & $|\getattr{\attribute}{\entity}|$ & Sample Values & Sample Prompts \\ 
 \midrule
\textbf{City} \\
Country & 158 & \makecell[{{p{1\linewidth}}}]{United States, China, Russia, Brazil, Australia} & \makecell[{{p{1\linewidth}}}]{\Verb|city to country: Toronto is in Canada. \{E\} is in|, \Verb|[\{"city": "Paris", "country": "France"\}, \{"city": "\{E\}",| \\ \Verb|\quad\quad "country": "|}\\
Continent & 6 & \makecell[{{p{1\linewidth}}}]{Asia, Europe, Africa, North America, South America} & \makecell[{{p{1\linewidth}}}]{\Verb|\{E\} is a city in the continent of|, \\ \Verb|[\{"city": "\{E\}", "continent": "|}\\
Latitude & 122 & \makecell[{{p{1\linewidth}}}]{41, 37, 47, 36, 35} & \makecell[{{p{1\linewidth}}}]{\Verb|[\{"city": "Rio de Janeiro", "lat": "23"\}, | \\ \Verb|\quad\quad \{"city": "\{E\}", "lat": "|, \Verb|[\{"city": "\{E\}", "lat": "|}\\
Longitude & 317 & \makecell[{{p{1\linewidth}}}]{30, 9, 10, 33, 11} & \makecell[{{p{1\linewidth}}}]{\Verb|[\{"city": "Rome", "long": "12.5"\}, \{"city": "\{E\}", "long": "|, \Verb| "long": "122.4"\}, \{"city": "\{E\}", "long": "|}\\
Language & 159 & \makecell[{{p{1\linewidth}}}]{English, Spanish, Chinese, Russian, Portuguese} & \makecell[{{p{1\linewidth}}}]{\Verb|[\{"city": "Beijing", "lang": "Chinese"\}, | \\ \Verb|\quad\quad \{"city": "\{E\}", "lang": "|, \\ \Verb|[\{"city": "\{E\}", "official language": "|}\\
Timezone & 267 & \makecell[{{p{1\linewidth}}}]{America/Chicago, Asia/Shanghai, Asia/Kolkata, \\ Europe/Moscow, America/Sao\_Paulo} & \makecell[{{p{1\linewidth}}}]{\Verb|Time zone in Los Angeles is America/Santiago; | \\ \Verb|\quad\quad Time zone in \{E\} is|, \\ \Verb|[\{"city": "New Delhi", "timezone": "UTC+5:30"\},| \\ \Verb|\quad\quad\{"city": "\{E\}", "timezone": "UTC|}\\
\\
\multicolumn{3}{l}{\textbf{Nobel Laureate}} \\
Field & 7 & \makecell[{{p{1\linewidth}}}]{Medicine, Physics, Chemistry, Literature, Peace} & \makecell[{{p{1\linewidth}}}]{\Verb|Jules A. Hoffmann won the Nobel Prize in Medicine.| \\ \Verb|\quad\quad\{E\} won the Nobel Prize in|, \\
\Verb|name: \{E\}, award: Nobel Prize in|} \\
Award Year & 118 & \makecell[{{p{1\linewidth}}}]{2001, 2019, 2009, 2011, 2000} & \makecell[{{p{1\linewidth}}}]{\Verb|"name": \{E\},  "award": "Nobel Prize", "year": "|, \\
\Verb|laureate: Frances H. Arnold, year: 2018, laureate: \{E\}, year:|}\\
Birth Year & 145 & \makecell[{{p{1\linewidth}}}]{1918, 1940, 1943, 1911, 1941} & \makecell[{{p{1\linewidth}}}]{\Verb|Alan Heeger was born in 1936. \{E\} was born in|, \\
\Verb|laureate: \{E\}, date of birth (YYYY-MM-DD):|}\\
Country of Birth & 81 & \makecell[{{p{1\linewidth}}}]{United States, United Kingdom, Germany, France, Sweden} & \makecell[{{p{1\linewidth}}}]{\Verb|name: \{E\}, country:|, \\
\Verb|Roderick MacKinnon was born in United States. \{E\} was born in|}\\
Gender & 4 & \makecell[{{p{1\linewidth}}}]{his, male, female, her} & \makecell[{{p{1\linewidth}}}]{\Verb|name: \{E\}, gender:|, \\
\Verb|David M. Lee: for his contributions in physics. \{E\}: for|}\\

 \bottomrule
\end{tabular}
 \caption{Attributes in \ourdataset. $|\getattr{\attribute}{\entity}|$ is the number of unique attribute values. In sampled prompts, \texttt{\{E\}} is a placeholder for the entity.}
 \vspace{-3ex}
 \label{tab:dataset_attribute}
\end{table*}

\begin{table*}[t]
\ContinuedFloat
  \centering
  \small
  \begin{tabular}{p{0.07\linewidth}p{0.03\linewidth}p{0.2\linewidth}p{0.6\linewidth}}
 \toprule
  Attributes & $|\getattr{\attribute}{\entity}|$ & Sample Values & Sample Prompts \\ 
 \midrule
\textbf{Verb} \\
Definition & 986 & \makecell[{{p{1\linewidth}}}]{take hold of, make certain, show, express in words, make} & \makecell[{{p{1\linewidth}}}]{\Verb|talk: communicate by speaking; win: achieve victory; \{E\}:|, \Verb|like: have a positive preference; walk: move on foot; \{E\}:|}\\
Past Tense & 986 & \makecell[{{p{1\linewidth}}}]{expanded, sealed, terminated, escaped, answered} & \makecell[{{p{1\linewidth}}}]{\Verb|present tense: \{E\}, past tense:|, \\ \Verb|write: wrote; look: looked; \{E\}:|}\\
\makecell[{{p{1\linewidth}}}]{Pronun-\\ciation} & 986 & \makecell[{{p{1\linewidth}}}]{\textipa{k@n"fju:z}, \textipa{fI"nIS}, \textipa{bOIl}, \textipa{In"SU@r}, \textipa{tIp}} & \makecell[{{p{1\linewidth}}}]{\Verb|create: \textipa{kri"eIt}; become: \textipa{bI"k\textturnv m}; \{E\}:|, \\ \Verb|begin: \textipa{bI"gIn}; change: \textipa{tSeIndZ}; \{E\}:|}\\
Singular & 986 & \makecell[{{p{1\linewidth}}}]{compensates, kicks, hunts, earns, accompanies} & \makecell[{{p{1\linewidth}}}]{\Verb|tell: tells; create: creates; \{E\}:|, \\ \Verb|present tense: \{E\}, 3rd person present:|}\\
\\
\multicolumn{3}{l}{\textbf{Physical Object}} \\
Category & 29 & \makecell[{{p{1\linewidth}}}]{plant, non-living thing, animal, NO, fruit} & \makecell[{{p{1\linewidth}}}]{
\Verb|bird is a type of animal: YES; rock is a type of animal: NO; |\\\Verb|\quad\quad\{E\} is a type of animal:|, \\
\Verb|Among the categories "plant", "animal", and "non-living thing",|\\\Verb|\quad\quad \{E\} belongs to "|}\\
Color & 12 & \makecell[{{p{1\linewidth}}}]{green, white, yellow, brown, black} & \makecell[{{p{1\linewidth}}}]{\Verb|The color of apple is usually red. The color of violet is|\\\Verb|\quad\quad usually purple. The color of \{E\} is usually|, \\
\Verb|The color of apple is usually red. The color of turquoise |\\\Verb|\quad\quad is usually blue. The color of \{E\} is usually|}\\
Size & 4 & \makecell[{{p{1\linewidth}}}]{cm, mm, m, km} & \makecell[{{p{1\linewidth}}}]{
\Verb|Among the units "mm", "cm", "m", and "km", | \\
\Verb|\quad\quad the size of \{E\} is usually on the scale of "|, \\
\Verb|Given the units "mm" "cm" "m" and "km",| \\ \Verb|\quad\quad the size of \{E\} usually is in "|} \\
Texture & 2 & \makecell[{{p{1\linewidth}}}]{soft, hard} & \makecell[{{p{1\linewidth}}}]{\Verb| hard or soft: rock is hard; towel is soft; |\\\Verb|\quad\quad blackberry is soft; wood is hard; \{E\} is|, \\
\Verb|Texture: rock: hard; towel: soft; blackberry: soft; |\\\Verb|\quad\quad charcoal: hard; \{E\}:|}\\
\\
\textbf{Occupation} \\
Duty & 650 & \makecell[{{p{1\linewidth}}}]{treat patients, teach students, sell products, create art, serve food} & \makecell[{{p{1\linewidth}}}]{\Verb|"occupation": "photographer", "duties": "to capture | \\\Verb|\quad\quad images using cameras"; "occupation": "\{E\}", "duties": "to|, \\
\Verb|"occupation": "\{E\}", "primary duties": "to|}\\
Gender Bias & 9 & \makecell[{{p{1\linewidth}}}]{he, male, his, female, she} & \makecell[{{p{1\linewidth}}}]{
\Verb|The \{E\} left early because| \\
\Verb|The newspaper praised the \{E\} for|}\\
Industry & 280 & \makecell[{{p{1\linewidth}}}]{construction, automotive, education, health care, agriculture} & \makecell[{{p{1\linewidth}}}]{\Verb|"occupation": "sales manager", "industry": "retail"; |\\ \Verb|\quad\quad "occupation": "\{E\}", "industry": "|, \\
\Verb|"occupation": "software developer",  "industry": |\\\Verb|\quad\quad "technology"; "occupation": "\{E\}", "industry": "|}\\
\makecell[{{p{1\linewidth}}}]{Work Location} & 128 & \makecell[{{p{1\linewidth}}}]{office, factory, hospital, construction site, studio} & \makecell[{{p{1\linewidth}}}]{\Verb|"occupation": "software developer", "environment": "office";|\\\Verb|\quad\quad  "occupation": "\{E\}", "environment": "|}\\

 \bottomrule
\end{tabular}
 \caption{Attributes in \ourdataset, continued.}
 \label{tab:dataset_attribute2}
\end{table*}

\subsection{Details of Entities and Attributes}
\label{appx:attribute}
We first identify entity types from existing datasets such as the Relations Dataset \cite{hernandez2023linearity} and \textsc{RippleEdits} \cite{cohen2023evaluating}, where each entity type potentially contains thousands of instances. We then source the entities and ground truth references for attribute values from online sources.\footnote{\url{https://github.com/kevinroberts/city-timezones}}
\footnote{\url{https://github.com/open-dict-data/ipa-dict/blob/master/data/en_US.txt}}
\footnote{\url{https://github.com/monolithpl/verb.forms.dictionary}}
\footnote{\url{https://www.nobelprize.org/prizes/lists/all-nobel-prizes/}}
\footnote{\url{https://huggingface.co/datasets/corypaik/coda}}
\footnote{\url{https://www.bls.gov/ooh,https://www.bls.gov/cps}} These online sources are distributed under MIT, Apache-2.0, and CC-0 licenses. Compared with similar entity types in the Relations Dataset and \textsc{RippleEdits}, \ourdataset\ has expanded the number of entities by a factor of at least 10 and included multiple attributes per entity.

We show the cardinality of the attributes, most frequent attribute values, and random samples of prompt templates in Table~\ref{tab:dataset_attribute}.

\subsection{The \ourdataset\  \llama-7B Instance}
\label{appx:dataset_llama2}
\begin{table*}[t!]
  \centering
  \small
  \begin{tabular}[b]{lccc@{}c}
 \toprule
  Entity Type & \#~Entities & \makecell[{{p{0.07\linewidth}}}]{\#~Prompts \\ Templates} & \makecell[{{p{0.2\linewidth}}}]{\#~Test Examples in \\ \quad$\entitysplit$/$\contextsplit$} & Accuracy (\%)\\ 
 \midrule
City & 800 & 90 & 15K/33K & 97.1 \\
Nobel Laureate  & 600 & 60 & 9K/23K & 94.3 \\
Verb & 600 & 40 & 12K/20K & 95.1 \\ 
Physical Object & 400 & 40 & 4K/6K & 94.3 \\
Occupation & 400 & 30 & 10K/16K & 96.4 \\
 \bottomrule
\end{tabular}
 \caption{Stats of \ourdataset\ in its \llama-7B instance, created by sampling a subset of examples where \llama-7B has a high accuracy in predicting attribute values.}
 \label{tab:dataset_appendix}
 \vspace{-3ex}
\end{table*}

The \ourdataset\ \llama-7B instance is used for benchmarking interpretability methods in Section~\ref{sec:results}. There are a total of 2800 entities in the \llama-7B instance. Table~\ref{tab:dataset_appendix} shows the number of entities, prompt templates, and test examples, i.e., the number of base--source input pairs for interchange intervention in the \llama-7B instance.

The \ourdataset\ \llama-7B instance is created by filtering examples where the pre-trained \llama-7B has a high accuracy in predicting attribute values. For each entity type, we take the $k$ entities with the highest accuracy over all prompt templates and the $n$ prompt templates with the highest accuracy over all entities, with the average accuracy over all prompts shown in Table~\ref{tab:dataset_appendix}. %
For most attributes, we directly compare model outputs against the ground truth attribute values. For ``latitude'' and ``longitude'' of a city, we relax the match to be $\pm$2 within the ground truth value. For ``pronunciation'' of a verb, we relax the match to allow variations in the transcription. For attributes with more open-ended outputs, including ``definition'' of a verb and ``duty'' of an occupation, we manually verify if the outputs are sensible. For ``gender bias'' of an occupation, we check for the consistency of gender bias over a set of prompts that instruct the model to output gender pronouns.

\section{Method Details}
\label{appx:method}

\subsection{PCA}

For PCA, we extract the 4096-dimension hidden state representations at the target intervention site as the inputs. The representations are first normalized to zero-mean and unit-variance using mean and variance estimated from the training set. We use the \texttt{sklearn} implementation\footnote{\url{https://scikit-learn.org/stable/modules/generated/sklearn.decomposition.PCA.html}} to compute the principal components. We then apply L1-based feature selection\footnote{\url{https://scikit-learn.org/stable/modules/feature_selection.html\#l1-based-feature-selection}} to identify a set of dimensions that most likely encode the target attribute $\attribute$. We undo the normalization after projecting back to the original space. 

We vary the coefficient of the L1 penalty, i.e., the parameter ``C'' in the \texttt{sklearn} implementation, to experiment with different intervention dimensions. We experiment with C $\in\{0.1, 1, 10, 1000\}$. We observe that regardless of the intervention dimension, the features selected have a high overlap with the first~$k$ principal components. For most attributes, the highest $\Score$ score is achieved when using the largest intervention dimension.

\subsection{Sparse Autoencoder}

For the sparse autoencoder, we use a single-layer encoder-decoder model.\footnote{\url{https://colab.research.google.com/drive/1u8larhpxy8w4mMsJiSBddNOzFGj7_RTn?usp=sharing\#scrollTo=Kn1E_44gCa-Z}} The autoencoder is trained on Wikipedia data as described below.

\paragraph{Model} Encoder: Fully connected layer with ReLU activations, dimensions $4096\times16384$. Decoder: Fully connected layer, dimensions $16384\times4096$. Latent dimension: $4\times4096$. The model is trained to optimize a combination of an L2 loss to reconstruct the representation and an L1 loss to enforce sparsity. 

\paragraph{Training Data} For each entity type, we sample 100k sentences from the Wikipedia corpus, each containing a mention of an entity in the training set. We extract the 4096-dimension hidden states at the target intervention site as the input for training the sparse autoencoder.

Similar to PCA, we apply L1-based feature selection on the latent representation to identify a set of dimensions that most likely encode the target attribute $\attribute$. We vary the coefficient C of the L1 penalty to experiment with different intervention dimension. The optimal C varies across attributes.

\subsection{RLAP}

\begin{table*}[t]
  \centering
  \small
  \begin{tabular}{p{0.15\linewidth}p{0.1\linewidth}p{0.1\linewidth}}
 \toprule
Attribute & $\entitysplit$ & $\contextsplit$ \\
\midrule
\multicolumn{2}{l}{City} \\
Country & 0.78 & 1.00 \\
Continent & 0.96 & 1.00 \\
Latitude & 0.18 & 1.00 \\
Longitude & 0.13 & 1.00 \\
Language & 0.60 & 1.00 \\
Timezone & 0.68 & 1.00 \\

\multicolumn{2}{l}{Nobel Laureate} \\
Field & 0.82 & 1.00 \\
Award Year & 0.08 & 1.00 \\
Birth Year & 0.01 & 1.00 \\
Country of Birth & 0.63 & 1.00 \\
Gender & 0.93 & 1.00 \\

\multicolumn{2}{l}{Verb} \\
Definition & 0.03 & 1.00 \\
Past Tense & 0.00 & 1.00 \\
Pronunciation & 0.00 & 1.00 \\
Singular & 0.00 & 1.00 \\

\multicolumn{2}{l}{Physical Object} \\
Category & 0.90 & 1.00 \\
Color & 0.49 & 1.00 \\
Size & 0.86 & 1.00 \\
Texture & 0.75 & 1.00 \\

\multicolumn{2}{l}{Occupation} \\
Duty & 0.06 & 1.00 \\
Gender Bias & 0.17 & 0.99 \\
Industry & 0.43 & 1.00 \\
Work Location & 0.44 & 1.00 \\

 \bottomrule
\end{tabular}
 \caption{Accuracy of linear probes on dev splits using the \llama-7B residual stream representations extracted from layer 7 above the last entity token. For most attribute, there exists a linear classifier with significant higher accuracy than random baseline on the entity dev split. For all attributes, there exists a linear classifier with close to perfect accuracy on the context dev split.}
 \label{tab:linear_classifier_results}
 \vspace{-3ex}
\end{table*}

RLAP learns a set of linear probes to find the feature $\feature$. Each linear probe aims to predict the attribute value from the entity representations. Similar to PCA and sparse autoencoders, we use the 4096-dimension hidden state representations at the target intervention site as the initial inputs and the corresponding attribute value as labels. In the case of attributes with extremely large output spaces, e.g., numerical outputs, we approximate the output with the first token. Table~\ref{tab:linear_classifier_results} shows the linear classifier accuracy on each attribute classification task.

We use the official R-LACE implementation\footnote{\url{https://github.com/shauli-ravfogel/rlace-icml}} and extract the rank-k orthogonal matrix $W$ from the final null projection\footnote{\url{https://github.com/shauli-ravfogel/rlace-icml/blob/master/rlace.py\#L90}} as $\feature_{\attribute}$. For each attribute, we experiment with rank $k\in\{32, 128, 512, 2048\}$. We run the algorithm for 100 iterations and select the rank with the highest $\Score$ score on the dev set. The optimal intervention dimension is usually small, i.e., 32 or 128, for attributes that have a high accuracy linear classifier.

\subsection{DBM-based and DAS-based Methods}
\label{appx:das}
For DBM- and DAS-based methods, we use the implementation from the \texttt{pyvene} library.\footnote{\url{https://github.com/stanfordnlp/pyvene}} For training data, both methods are trained on base--source pairs with interchange interventions.

For DBM and MDBM, we use a starting temperature of $1e{-}2$ and gradually reducing it to $1e{-}7$. The feature dimension is controlled by the coefficient of the L1 loss. The optimal coefficient for the DBM penalty is around 0.001, while no penalty generally works better for MDBM, as the multi-task objective naturally encourages the methods to select as few dimensions as possible.

For DAS and MDAS, we do not instantiate the full rotation matrix, but only parameterize the $k$ orthogonal vectors that form the feature $\feature_{\attribute}$. The interchange intervention is defined as
\[
\II(\model, \feature_{\attribute}, x, x')= (I - W^{\top}W)(\GetVals(\model(x), \rep)) + W^{\top}W(\GetVals(\model(x'), \rep))
\]
where the rows of $W$ are the $k$ orthogonal vectors.
We experiment with $k\in\{32, 128, 512, 2048\}$ and select the dimension with the highest $\Score$ score on the dev set. For most attributes, a larger intervention dimension, e.g., 512 or 2048, leads to a higher $\Score$ score.

\subsection{Computational Cost}

All models are trained and evaluated on a single NVIDIA RTX A6000 GPU.

For training, the computational cost of sparse autoencoders is the lowest, as training sparse autoencoders does not involve backpropagating through the original \llama-7B model or computing orthogonal factorization of weight matrices. Each epoch of the sparse autoencoder training, i.e., iterating over 100k examples, takes about 100 seconds with \llama-7B features extracted offline. The computational cost of RLAP- and DAS-based method largely depends on the rank of the nullspace or the intervention dimension, i.e., the number of orthogonal vectors. For RLAP, it takes 1 hour per 100 iterations with a feature dimension 4096 and a target rank of 128. For DAS and MDAS with the reduced parameter formulation, the training time for an intervention dimension of 128 (out of a feature dimension of 4096) over 1k intervention examples is about 50 seconds. The computational cost of DBM-based method is about 35 seconds per 1k intervention examples. 

For evaluation, the inference speed of our proposed framework is 20 seconds per 1k intervention examples.

\section{Results}
\label{sec:results_appx}
\begin{table*}[tp]

\begin{subtable}[t]{1.0\textwidth}
  \centering
  \small
  \begin{tabular}[b]{lc@{\hspace{3px}}cc@{\hspace{3px}}cc@{\hspace{3px}}cc@{\hspace{3px}}cc@{\hspace{3px}}cc@{\hspace{3px}}cc@{\hspace{3px}}cc@{\hspace{3px}}cc@{\hspace{3px}}cc@{\hspace{3px}}cc@{\hspace{3px}}c|c@{\hspace{3px}}cc@{\hspace{3px}}cc@{\hspace{3px}}cc@{\hspace{3px}}cc@{\hspace{3px}}cc@{\hspace{3px}}cc@{\hspace{3px}}cc@{\hspace{3px}}cc@{\hspace{3px}}c|c@{\hspace{3px}}cc@{\hspace{3px}}cc@{\hspace{3px}}cc@{\hspace{3px}}cc@{\hspace{3px}}cc@{\hspace{3px}}cc@{\hspace{3px}}cc@{\hspace{3px}}cc@{\hspace{3px}}cc@{\hspace{3px}}c|c@{\hspace{3px}}cc@{\hspace{3px}}cc@{\hspace{3px}}cc@{\hspace{3px}}cc@{\hspace{3px}}cc@{\hspace{3px}}cc@{\hspace{3px}}cc@{\hspace{3px}}cc@{\hspace{3px}}cc@{\hspace{3px}}c|c@{\hspace{3px}}cc@{\hspace{3px}}cc@{\hspace{3px}}cc@{\hspace{3px}}cc@{\hspace{3px}}cc@{\hspace{3px}}cc@{\hspace{3px}}cc@{\hspace{3px}}cc@{\hspace{3px}}cc@{\hspace{3px}}cc@{\hspace{3px}}c|c@{\hspace{3px}}cc@{\hspace{3px}}cc@{\hspace{3px}}cc@{\hspace{3px}}cc@{\hspace{3px}}cc@{\hspace{3px}}cc@{\hspace{3px}}cc@{\hspace{3px}}cc@{\hspace{3px}}cc@{\hspace{3px}}c}
 \toprule
  Method & \multicolumn{2}{c}{Continent} & \multicolumn{2}{c}{Country} & \multicolumn{2}{c}{Language} & \multicolumn{2}{c}{Latitude} & \multicolumn{2}{c}{Longitude} & \multicolumn{2}{c}{Timezone} & \nochangeScore & \changeScore & \Score \\
 \midrule
 \entitysplit \\
 PCA & 32.7 & 45.2 & 36.3 & 58.6 & 34.2 & 33.3 & 32.7 & 44.2 & 39.3 & 35.4 & 36.0 & 36.6 & 35.2 & 42.2 & 38.7 \\
SAE & 82.5 & 15.2 & 40.4 & 70.0 & 91.8 & 5.0 & 92.1 & 17.4 & 93.3 & 21.2 & 91.1 & 13.6 & 81.9 & 23.7 & 52.8 \\
RLAP & 89.4 & 21.0 & 38.2 & 55.8 & 44.6 & 48.0 & 58.1 & 48.2 & 38.2 & 54.0 & 41.1 & 50.0 & 51.6 & 46.2 & 48.9 \\
DBM & 65.9 & 70.0 & 44.8 & 70.6 & 42.9 & 54.3 & 45.1 & 59.8 & 44.9 & 57.0 & 72.2 & 54.0 & 52.6 & 61.0 & 56.8 \\
DAS & 67.3 & 86.4 & 30.1 & 83.8 & 36.3 & 74.0 & 52.7 & 63.2 & 50.3 & 56.6 & 71.0 & 74.0 & 51.3 & 73.0 & 62.1 \\
MDBM & 72.6 & 68.2 & 58.6 & 73.0 & 56.7 & 52.3 & 59.1 & 55.2 & 59.9 & 54.4 & 75.7 & 56.4 & 63.8 & 59.9 & 61.8 \\
MDAS & 92.1 & 69.2 & 82.7 & 65.6 & 86.4 & 51.7 & 91.4 & 47.6 & 93.1 & 46.0 & 92.9 & 62.4 & 89.8 & 57.1 & 73.4 \\
 \contextsplit \\
 PCA & 27.9 & 46.1 & 31.4 & 52.5 & 29.2 & 19.0 & 26.8 & 40.0 & 27.5 & 53.0 & 28.8 & 47.5 & 28.6 & 43.0 & 35.8 \\
SAE & 65.6 & 28.9 & 29.3 & 75.4 & 88.6 & 4.5 & 87.0 & 18.0 & 88.4 & 26.5 & 65.8 & 27.0 & 70.8 & 30.0 & 50.4 \\
RLAP & 86.0 & 21.4 & 22.4 & 84.7 & 36.8 & 43.0 & 46.1 & 55.0 & 28.3 & 72.5 & 34.8 & 51.0 & 42.4 & 54.6 & 48.5 \\
DBM & 58.7 & 58.6 & 37.9 & 66.0 & 36.4 & 36.0 & 38.3 & 61.4 & 38.9 & 69.0 & 67.4 & 53.5 & 46.3 & 57.4 & 51.8 \\
DAS & 58.9 & 84.9 & 17.7 & 89.3 & 27.7 & 54.0 & 33.9 & 77.6 & 40.9 & 72.5 & 64.6 & 73.5 & 40.6 & 75.3 & 58.0 \\
MDBM & 65.4 & 56.4 & 50.7 & 67.6 & 52.1 & 32.0 & 51.9 & 58.2 & 53.3 & 66.5 & 70.0 & 55.5 & 57.2 & 56.0 & 56.6 \\
MDAS & 86.6 & 64.9 & 70.5 & 70.7 & 90.3 & 20.0 & 88.0 & 57.0 & 89.8 & 62.0 & 90.0 & 57.5 & 85.9 & 55.4 & 70.6 \\
 \bottomrule
\end{tabular}
 \caption{Scores of city attributes.}
 \label{tab:results_city}
\end{subtable}

\begin{subtable}[t]{1.0\textwidth}
  \centering
  \small
  \begin{tabular}[b]{lc@{\hspace{3px}}cc@{\hspace{3px}}cc@{\hspace{3px}}cc@{\hspace{3px}}cc@{\hspace{3px}}cc@{\hspace{3px}}cc@{\hspace{3px}}cc@{\hspace{3px}}cc@{\hspace{3px}}cc@{\hspace{3px}}cc@{\hspace{3px}}cc@{\hspace{3px}}c|c@{\hspace{3px}}cc@{\hspace{3px}}cc@{\hspace{3px}}cc@{\hspace{3px}}cc@{\hspace{3px}}cc@{\hspace{3px}}cc@{\hspace{3px}}cc@{\hspace{3px}}cc@{\hspace{3px}}cc@{\hspace{3px}}cc@{\hspace{3px}}cc@{\hspace{3px}}c|c@{\hspace{3px}}cc@{\hspace{3px}}cc@{\hspace{3px}}cc@{\hspace{3px}}cc@{\hspace{3px}}cc@{\hspace{3px}}cc@{\hspace{3px}}cc@{\hspace{3px}}cc@{\hspace{3px}}cc@{\hspace{3px}}cc@{\hspace{3px}}cc@{\hspace{3px}}cc@{\hspace{3px}}cc@{\hspace{3px}}cc@{\hspace{3px}}cc@{\hspace{3px}}cc@{\hspace{3px}}cc@{\hspace{3px}}c|c@{\hspace{3px}}cc@{\hspace{3px}}cc@{\hspace{3px}}cc@{\hspace{3px}}cc@{\hspace{3px}}cc@{\hspace{3px}}cc@{\hspace{3px}}c|c@{\hspace{3px}}cc@{\hspace{3px}}cc@{\hspace{3px}}cc@{\hspace{3px}}cc@{\hspace{3px}}cc@{\hspace{3px}}cc@{\hspace{3px}}cc@{\hspace{3px}}c}
 \toprule
  Method & \multicolumn{2}{c}{Award Year} & \multicolumn{2}{c}{Birth Year} & \multicolumn{2}{c}{Country of Birth} & \multicolumn{2}{c}{Field} & \multicolumn{2}{c}{Gender} & \nochangeScore & \changeScore & \Score \\
 \midrule
 \entitysplit \\
 PCA & 24.2 & 22.7 & 30.8 & 2.3 & 22.4 & 70.0 & 24.3 & 78.3 & 4.3 & 81.0 & 21.2 & 50.9 & 36.0 \\
SAE & 79.8 & 0.7 & 80.1 & 0.7 & 39.8 & 49.0 & 43.4 & 54.0 & 71.3 & 63.7 & 62.9 & 33.6 & 48.2 \\
RLAP & 87.3 & 0.3 & 90.3 & 1.0 & 68.0 & 8.7 & 82.5 & 54.0 & 95.3 & 71.0 & 84.7 & 27.0 & 55.8 \\
DBM & 91.8 & 0.7 & 98.6 & 0.3 & 61.5 & 32.0 & 71.3 & 57.7 & 92.6 & 71.7 & 83.2 & 32.5 & 57.8 \\
DAS & 57.1 & 5.0 & 72.7 & 2.3 & 80.9 & 25.3 & 80.1 & 72.7 & 80.8 & 77.7 & 74.3 & 36.6 & 55.5 \\
MDBM & 40.8 & 19.3 & 70.2 & 2.0 & 66.9 & 36.3 & 69.2 & 62.3 & 76.4 & 79.7 & 64.7 & 39.9 & 52.3 \\
MDAS & 83.6 & 4.0 & 85.2 & 2.0 & 88.8 & 28.0 & 86.9 & 58.0 & 93.4 & 78.0 & 87.6 & 34.0 & 60.8 \\
 \contextsplit \\
 PCA & 19.2 & 25.4 & 22.6 & 3.3 & 18.4 & 73.2 & 23.6 & 76.0 & 3.0 & 67.0 & 17.4 & 49.0 & 33.2 \\
SAE & 74.9 & 1.0 & 73.8 & 1.0 & 38.1 & 38.3 & 65.1 & 28.0 & 64.8 & 35.0 & 63.3 & 20.7 & 42.0 \\
RLAP & 88.1 & 0.4 & 90.3 & 0.8 & 54.4 & 67.3 & 77.7 & 67.3 & 94.0 & 61.0 & 80.9 & 39.4 & 60.1 \\
DBM & 88.1 & 0.2 & 96.9 & 0.0 & 50.6 & 50.2 & 56.1 & 59.3 & 96.8 & 61.7 & 77.7 & 34.3 & 56.0 \\
DAS & 42.7 & 18.4 & 13.9 & 7.5 & 37.1 & 72.8 & 30.2 & 82.3 & 88.0 & 72.7 & 42.4 & 50.7 & 46.5 \\
MDBM & 38.6 & 20.6 & 69.5 & 2.2 & 65.8 & 54.2 & 66.7 & 65.7 & 91.6 & 72.0 & 66.4 & 42.9 & 54.7 \\
MDAS & 80.2 & 27.4 & 83.9 & 12.3 & 86.6 & 72.8 & 90.2 & 72.0 & 93.4 & 73.0 & 86.9 & 51.5 & 69.2 \\
 \bottomrule
\end{tabular}
 \caption{Scores of Nobel laureate attributes.}
 \label{tab:results_nobel_prize_winner}
\end{subtable}

\caption{Per-task results.}
\vspace{-3ex}
\end{table*}

\begin{table*}
    
\ContinuedFloat

\begin{subtable}[t]{1.0\textwidth}
  \centering
  \small
  \begin{tabular}[b]{lc@{\hspace{3px}}cc@{\hspace{3px}}cc@{\hspace{3px}}cc@{\hspace{3px}}cc@{\hspace{3px}}cc@{\hspace{3px}}cc@{\hspace{3px}}cc@{\hspace{3px}}cc@{\hspace{3px}}cc@{\hspace{3px}}cc@{\hspace{3px}}cc@{\hspace{3px}}c|c@{\hspace{3px}}cc@{\hspace{3px}}cc@{\hspace{3px}}cc@{\hspace{3px}}cc@{\hspace{3px}}cc@{\hspace{3px}}cc@{\hspace{3px}}cc@{\hspace{3px}}cc@{\hspace{3px}}cc@{\hspace{3px}}cc@{\hspace{3px}}cc@{\hspace{3px}}c|c@{\hspace{3px}}cc@{\hspace{3px}}cc@{\hspace{3px}}cc@{\hspace{3px}}cc@{\hspace{3px}}cc@{\hspace{3px}}cc@{\hspace{3px}}cc@{\hspace{3px}}cc@{\hspace{3px}}cc@{\hspace{3px}}cc@{\hspace{3px}}cc@{\hspace{3px}}cc@{\hspace{3px}}cc@{\hspace{3px}}cc@{\hspace{3px}}c|c@{\hspace{3px}}cc@{\hspace{3px}}cc@{\hspace{3px}}cc@{\hspace{3px}}cc@{\hspace{3px}}cc@{\hspace{3px}}cc@{\hspace{3px}}cc@{\hspace{3px}}cc@{\hspace{3px}}cc@{\hspace{3px}}c}
 \toprule
  Method & \multicolumn{2}{c}{Definition} & \multicolumn{2}{c}{Past Tense} & \multicolumn{2}{c}{Pronunciation} & \multicolumn{2}{c}{Singular} & \nochangeScore & \changeScore & \Score \\
 \midrule
 \entitysplit \\
 PCA & 4.9 & 59.5 & 4.6 & 95.3 & 2.1 & 66.5 & 4.2 & 93.3 & 4.0 & 78.6 & 41.3 \\
SAE & 93.4 & 3.5 & 15.4 & 87.3 & 85.4 & 3.0 & 14.3 & 82.3 & 52.1 & 44.0 & 48.1 \\
RLAP & 22.1 & 42.0 & 15.8 & 87.3 & 23.9 & 45.5 & 13.5 & 85.3 & 18.8 & 65.0 & 41.9 \\
DBM & 22.0 & 51.0 & 16.3 & 88.7 & 10.2 & 58.0 & 14.2 & 87.0 & 15.7 & 71.2 & 43.4 \\
DAS & 90.3 & 12.0 & 11.9 & 92.0 & 89.4 & 19.5 & 13.6 & 85.8 & 51.3 & 52.3 & 51.8 \\
MDBM & 55.8 & 30.0 & 32.8 & 70.5 & 66.4 & 20.0 & 25.4 & 75.8 & 45.1 & 49.1 & 47.1 \\
MDAS & 97.6 & 6.5 & 88.4 & 1.2 & 89.5 & 25.0 & 85.4 & 2.5 & 90.2 & 8.8 & 49.5 \\
 \contextsplit \\
 PCA & 9.6 & 57.0 & 8.3 & 84.3 & 4.3 & 44.0 & 9.2 & 78.3 & 7.9 & 65.9 & 36.9 \\
SAE & 84.3 & 10.5 & 16.8 & 77.3 & 74.1 & 5.5 & 16.2 & 73.7 & 47.9 & 41.8 & 44.8 \\
RLAP & 19.5 & 46.5 & 15.0 & 80.7 & 19.1 & 46.5 & 13.9 & 79.3 & 16.9 & 63.2 & 40.0 \\
DBM & 21.7 & 53.0 & 16.3 & 84.3 & 12.3 & 52.5 & 14.7 & 81.0 & 16.3 & 67.7 & 42.0 \\
DAS & 69.5 & 36.5 & 8.7 & 93.3 & 77.4 & 49.0 & 7.4 & 89.7 & 40.7 & 67.1 & 53.9 \\
MDBM & 64.4 & 29.5 & 28.4 & 70.0 & 62.9 & 28.0 & 27.5 & 68.0 & 45.8 & 48.9 & 47.3 \\
MDAS & 94.5 & 21.5 & 74.2 & 17.3 & 84.3 & 44.0 & 70.3 & 24.3 & 80.8 & 26.8 & 53.8 \\
 \bottomrule
\end{tabular}
 \caption{Scores of verb attributes.}
 \label{tab:results_verb}
\end{subtable}

\begin{subtable}[t]{1.0\textwidth}
  \centering
  \small
  \begin{tabular}[b]{lc@{\hspace{3px}}cc@{\hspace{3px}}cc@{\hspace{3px}}cc@{\hspace{3px}}cc@{\hspace{3px}}cc@{\hspace{3px}}cc@{\hspace{3px}}cc@{\hspace{3px}}cc@{\hspace{3px}}cc@{\hspace{3px}}c|c@{\hspace{3px}}cc@{\hspace{3px}}cc@{\hspace{3px}}cc@{\hspace{3px}}cc@{\hspace{3px}}cc@{\hspace{3px}}cc@{\hspace{3px}}c|c@{\hspace{3px}}cc@{\hspace{3px}}cc@{\hspace{3px}}cc@{\hspace{3px}}cc@{\hspace{3px}}cc@{\hspace{3px}}c|c@{\hspace{3px}}cc@{\hspace{3px}}cc@{\hspace{3px}}cc@{\hspace{3px}}cc@{\hspace{3px}}cc@{\hspace{3px}}cc@{\hspace{3px}}cc@{\hspace{3px}}cc@{\hspace{3px}}c}
 \toprule
  Method & \multicolumn{2}{c}{Category} & \multicolumn{2}{c}{Color} & \multicolumn{2}{c}{Size} & \multicolumn{2}{c}{Texture} & \nochangeScore & \changeScore & \Score \\
 \midrule
 \entitysplit \\
 PCA & 45.6 & 49.8 & 35.1 & 63.7 & 27.7 & 50.5 & 26.3 & 47.5 & 33.7 & 52.9 & 43.3 \\
SAE & 94.2 & 7.9 & 34.2 & 63.2 & 95.0 & 3.0 & 95.3 & 29.0 & 79.6 & 25.8 & 52.7 \\
RLAP & 85.6 & 30.6 & 83.9 & 8.0 & 62.0 & 28.5 & 58.7 & 47.5 & 72.5 & 28.7 & 50.6 \\
DBM & 70.1 & 35.6 & 62.0 & 40.0 & 98.0 & 2.0 & 97.7 & 30.0 & 81.9 & 26.9 & 54.4 \\
DAS & 77.3 & 52.0 & 79.7 & 28.7 & 87.2 & 24.0 & 92.0 & 47.5 & 84.0 & 38.1 & 61.1 \\
MDBM & 59.8 & 48.5 & 53.5 & 59.2 & 74.5 & 27.5 & 81.2 & 49.0 & 67.3 & 46.1 & 56.7 \\
MDAS & 85.1 & 49.8 & 87.0 & 19.8 & 88.5 & 19.5 & 91.5 & 46.5 & 88.0 & 33.9 & 60.9 \\
 \contextsplit \\
 PCA & 43.1 & 66.8 & 40.3 & 63.3 & 30.8 & 46.5 & 25.4 & 68.0 & 34.9 & 61.1 & 48.0 \\
SAE & 39.9 & 70.0 & 43.8 & 62.2 & 91.4 & 6.0 & 90.9 & 34.5 & 66.5 & 43.2 & 54.9 \\
RLAP & 83.6 & 47.2 & 82.3 & 22.5 & 64.6 & 30.0 & 60.9 & 61.0 & 72.8 & 40.2 & 56.5 \\
DBM & 72.1 & 47.2 & 64.6 & 46.0 & 97.3 & 2.5 & 97.5 & 32.5 & 82.9 & 32.1 & 57.5 \\
DAS & 70.7 & 75.8 & 72.2 & 67.8 & 82.2 & 53.5 & 85.6 & 64.5 & 77.7 & 65.4 & 71.5 \\
MDBM & 64.3 & 59.0 & 60.6 & 59.7 & 78.6 & 33.0 & 83.2 & 59.5 & 71.7 & 52.8 & 62.2 \\
MDAS & 84.8 & 73.0 & 83.1 & 61.5 & 87.8 & 46.0 & 86.3 & 65.0 & 85.5 & 61.4 & 73.4 \\
 \bottomrule
\end{tabular}
 \caption{Scores of physical object attributes.}
 \label{tab:results_physical_object}
\end{subtable}

\begin{subtable}[t]{1.0\textwidth}
  \centering
  \small
  \begin{tabular}[b]{lc@{\hspace{3px}}cc@{\hspace{3px}}cc@{\hspace{3px}}cc@{\hspace{3px}}cc@{\hspace{3px}}cc@{\hspace{3px}}c|c@{\hspace{3px}}cc@{\hspace{3px}}cc@{\hspace{3px}}cc@{\hspace{3px}}cc@{\hspace{3px}}cc@{\hspace{3px}}cc@{\hspace{3px}}cc@{\hspace{3px}}cc@{\hspace{3px}}cc@{\hspace{3px}}cc@{\hspace{3px}}cc@{\hspace{3px}}cc@{\hspace{3px}}c|c@{\hspace{3px}}cc@{\hspace{3px}}cc@{\hspace{3px}}cc@{\hspace{3px}}cc@{\hspace{3px}}cc@{\hspace{3px}}cc@{\hspace{3px}}cc@{\hspace{3px}}cc@{\hspace{3px}}cc@{\hspace{3px}}c|c@{\hspace{3px}}cc@{\hspace{3px}}cc@{\hspace{3px}}cc@{\hspace{3px}}cc@{\hspace{3px}}cc@{\hspace{3px}}cc@{\hspace{3px}}cc@{\hspace{3px}}cc@{\hspace{3px}}cc@{\hspace{3px}}cc@{\hspace{3px}}cc@{\hspace{3px}}cc@{\hspace{3px}}cc@{\hspace{3px}}cc@{\hspace{3px}}c}
 \toprule
  Method & \multicolumn{2}{c}{Duty} & \multicolumn{2}{c}{Gender Bias} & \multicolumn{2}{c}{Industry} & \multicolumn{2}{c}{Work Location} & \nochangeScore & \changeScore & \Score \\
 \midrule
 \entitysplit \\
 PCA & 39.9 & 33.7 & 28.1 & 61.7 & 36.3 & 38.0 & 35.9 & 31.0 & 35.1 & 41.1 & 38.1 \\
SAE & 68.9 & 4.0 & 57.1 & 49.0 & 61.7 & 10.5 & 64.3 & 13.0 & 63.0 & 19.1 & 41.1 \\
RLAP & 62.1 & 17.7 & 93.8 & 44.0 & 58.9 & 18.5 & 62.0 & 18.0 & 69.2 & 24.5 & 46.9 \\
DBM & 59.3 & 23.3 & 93.2 & 42.7 & 67.2 & 18.3 & 66.4 & 16.0 & 71.5 & 25.1 & 48.3 \\
DAS & 59.8 & 23.0 & 83.7 & 75.7 & 57.9 & 29.3 & 57.9 & 27.0 & 64.9 & 38.7 & 51.8 \\
MDBM & 52.0 & 35.3 & 81.7 & 66.0 & 57.8 & 29.5 & 59.3 & 24.5 & 62.7 & 38.8 & 50.8 \\
MDAS & 82.5 & 12.0 & 85.0 & 70.0 & 82.5 & 17.5 & 83.7 & 14.5 & 83.4 & 28.5 & 56.0 \\
 \contextsplit \\
 PCA & 39.2 & 45.0 & 21.9 & 68.0 & 33.8 & 42.7 & 38.3 & 44.5 & 33.3 & 50.0 & 41.7 \\
SAE & 66.7 & 7.7 & 47.7 & 61.0 & 58.9 & 14.3 & 65.1 & 14.5 & 59.6 & 24.4 & 42.0 \\
RLAP & 60.3 & 23.0 & 92.5 & 51.0 & 56.7 & 23.3 & 62.3 & 24.0 & 68.0 & 30.3 & 49.1 \\
DBM & 49.5 & 14.7 & 87.3 & 29.5 & 56.4 & 18.0 & 56.4 & 21.5 & 62.4 & 20.9 & 41.7 \\
DAS & 46.9 & 49.7 & 79.7 & 85.0 & 44.2 & 55.3 & 46.0 & 46.0 & 54.2 & 59.0 & 56.6 \\
MDBM & 43.6 & 22.7 & 77.7 & 70.5 & 54.2 & 31.3 & 60.9 & 27.0 & 59.1 & 37.9 & 48.5 \\
MDAS & 78.7 & 32.0 & 81.0 & 85.5 & 70.1 & 38.7 & 74.1 & 27.0 & 75.9 & 45.8 & 60.9 \\
 \bottomrule
\end{tabular}
 \caption{Scores of occupation attributes.}
 \label{tab:results_occupation}
\end{subtable}

\caption{Per-task results, continued.}
\label{tab:results_appendix}
\end{table*}

For all methods, we conduct hyper-parameter search on the dev set. We report a single-run test set results using the set of hyper-parameters that achieves the highest score on the dev set. For intervention site, we choose layer 16 for city attributes and layer 7 for the rest attributes.

\subsection{Breakdown of Benchmark Results}
Table~\ref{tab:results_appendix} shows the breakdown of benchmark results in Table~\ref{tab:results_all_methods}. For each method, we report a breakdown of the highest \Score\ score per attribute, i.e., the pair of \changeScore\ score and \nochangeScore\ score that add up to the highest \Score\ score. The final score in Table~\ref{tab:results_all_methods} is an average of the \Score\ score over all five entity types. For example, for PCA, the \Score\ score under the $\entitysplit$ setting is $(38.7+36.0+41.3+43.3+38.1)/5=39.5$.

\subsection{Additional Attribute Disentanglement Results}
\label{sec:more_heatmap}
\begin{figure*}[tp!]
     \centering
    \begin{subfigure}[t]{0.4\textwidth}
        \centering
        \includegraphics[height=30ex,trim={0 0 5cm 0},clip]{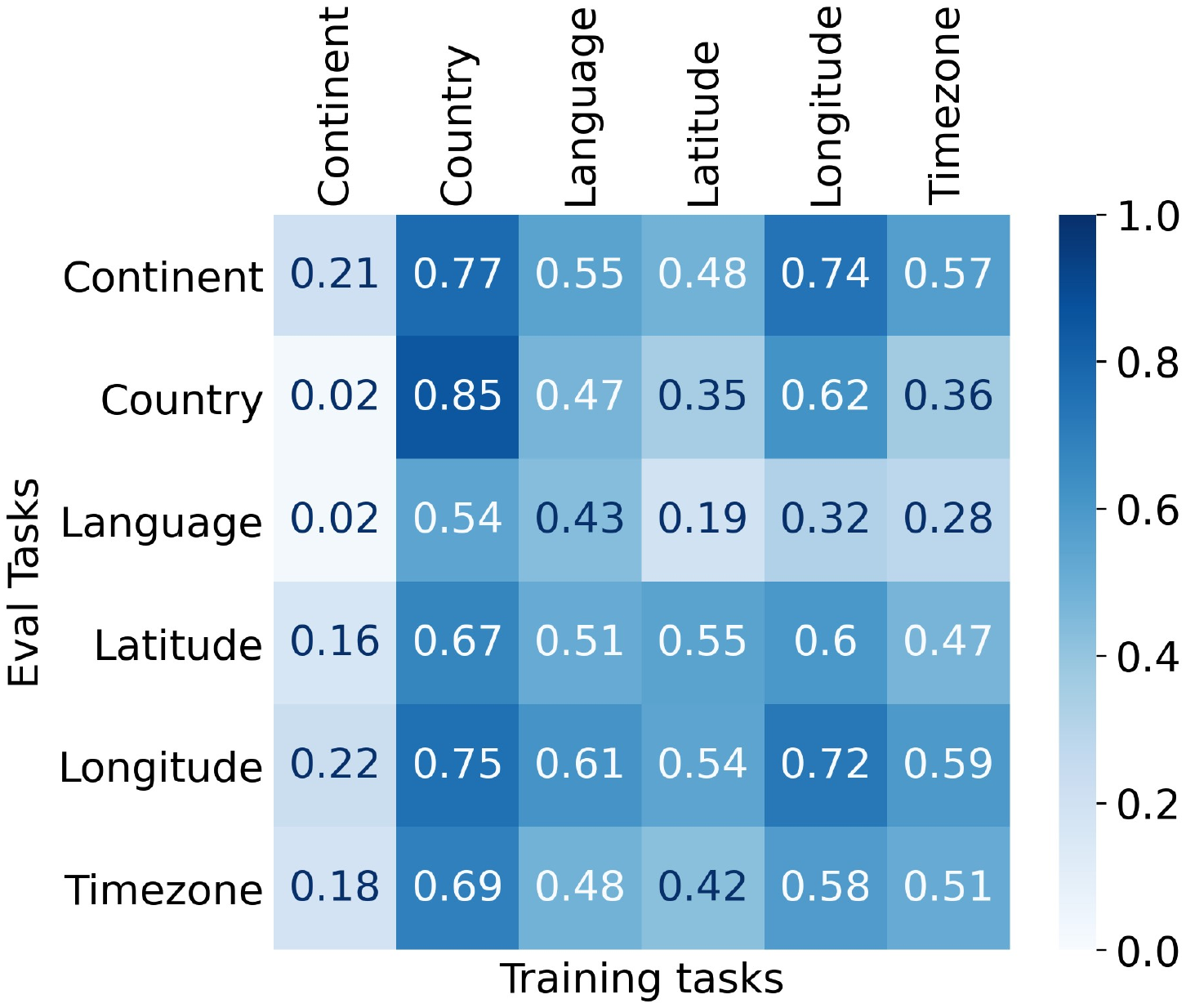}
        \caption{\changeScore\ score from RLAP.}
    \end{subfigure}%
    ~
    \begin{subfigure}[t]{0.27\textwidth}
         \includegraphics[height=30ex, trim={7.15cm 0 5cm 0},clip]{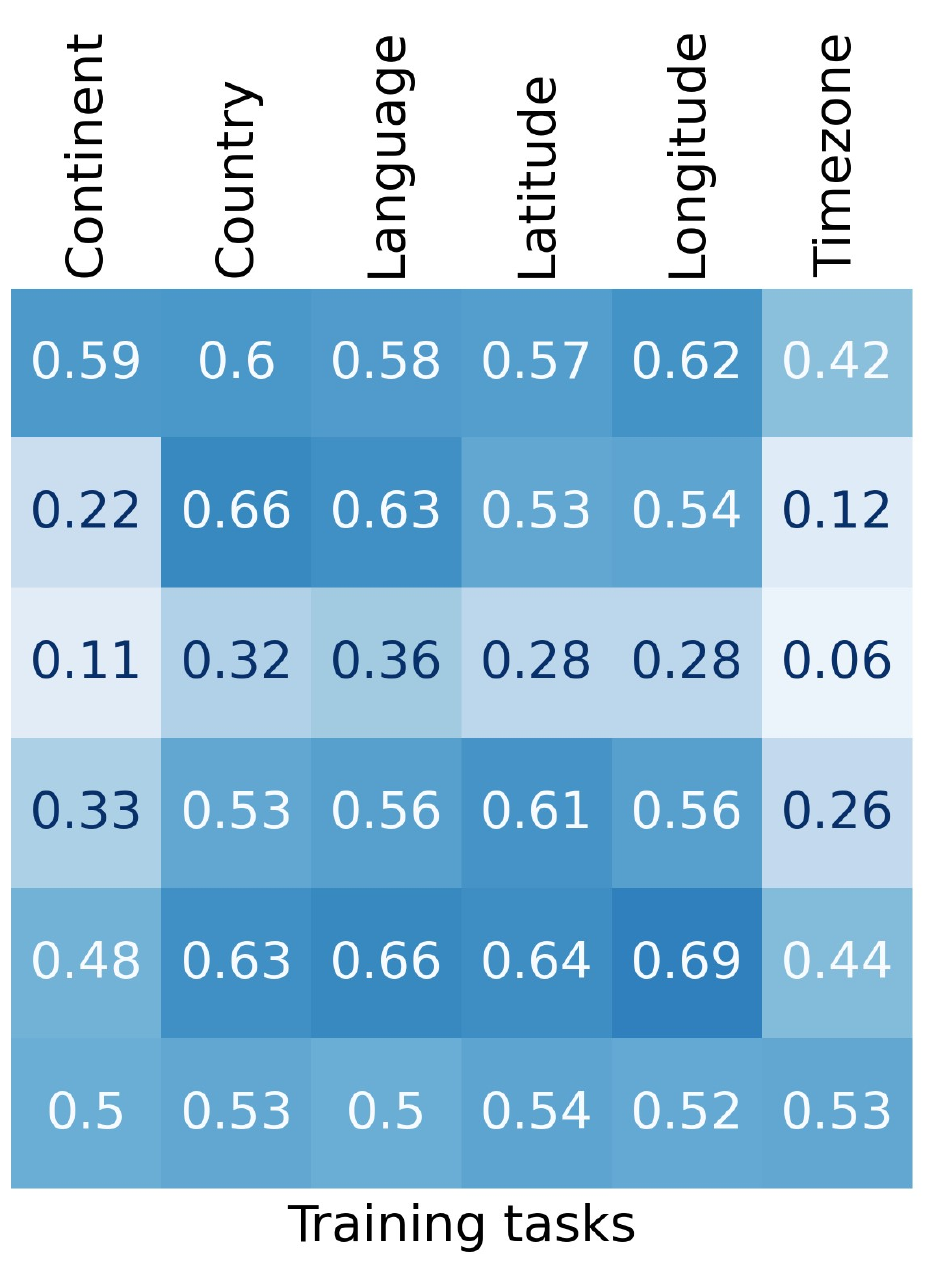}
        \caption{\changeScore\ score from DBM.}
    \end{subfigure}%
    \begin{subfigure}[t]{0.3\textwidth}
         \includegraphics[height=30ex, trim={7.15cm 0 5cm 0},clip]{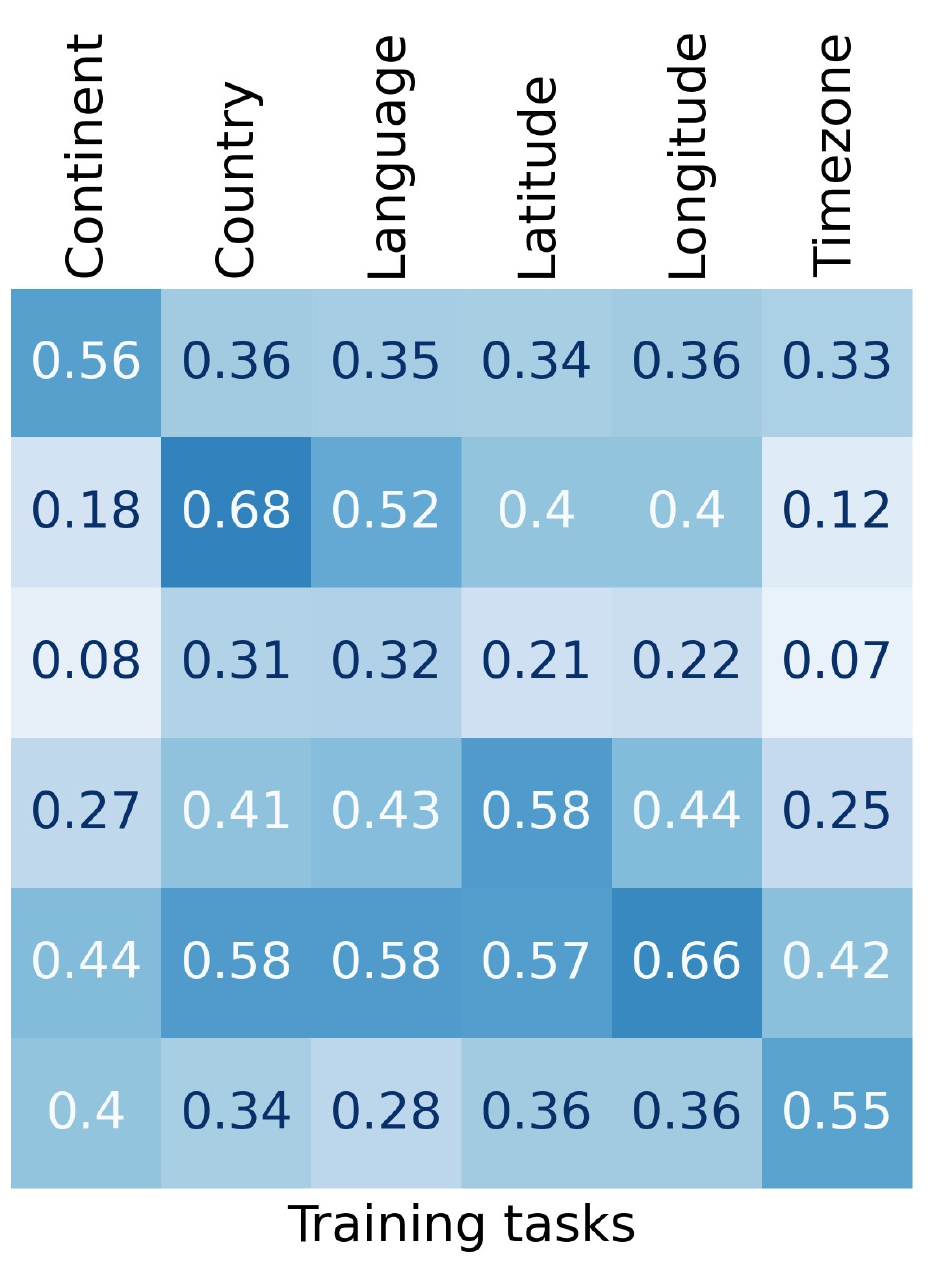}
        \caption{\changeScore\ score from MDBM.}
    \end{subfigure}
    \caption{Additional feature disentanglement results for RLAP, DBM, and MDBM methods.}
    \label{fig:heatmap_appendix}
\end{figure*}

In Figure~\ref{fig:moreresults}, we show the feature entanglement results from DAS and MDAS. We provide additional results from all other supervised methods: RLAP, DBM, and MDBM in Figure~\ref{fig:heatmap_appendix}. Though these methods are trained on different objectives and identify different features $\feature_{\attribute}$, they show similar patterns in terms of entanglement between attribute representations. For all methods, representations of most attributes are entangled with ``continent'' (and ``timezone'', which for most cases starts with the continent name). Representations of attributes such as ``county--language'' are also highly entangled.

\end{document}